\documentclass{article} 
\usepackage{iclr2024_conference,times}

\usepackage{hyperref}
\usepackage{url}
\usepackage{graphicx}
\usepackage{pifont}
\usepackage{wrapfig}
\usepackage{subcaption}
\usepackage{booktabs} 
\usepackage{multirow, mathtools} 
\usepackage{algorithm,algpseudocode}
\usepackage{dsfont}

\usepackage{amsmath,amsfonts,bm}

\def\eqref#1{(\ref{#1})}

\def\1{\bm{1}}

\DeclareMathAlphabet{\mathsfit}{\encodingdefault}{\sfdefault}{m}{sl}
\SetMathAlphabet{\mathsfit}{bold}{\encodingdefault}{\sfdefault}{bx}{n}

\DeclareMathOperator*{\argmin}{arg\,min}

\newcommand{\bdelta}{\boldsymbol{\delta}}
\newcommand{\btheta}{\boldsymbol{\theta}}
\newcommand{\CGE}{\text{CGE}}
\newcommand{\RGE}{\text{RGE}}
\newcommand{\DeepZero}{\text{DeepZero}}
\newcommand{\Def}[0]{\mathrel{\mathop:}=}
\newcommand{\grasp}{\text{GraSP}}
\newcommand{\zograsp}{\text{ZO-GraSP}}
\newcommand{\SR}{\text{SR}}

\newcommand{\txtr}{\textcolor{red}}
\newcommand{\txtg}{\textcolor{teal}}

\DeclareMathOperator*{\minimize}{\text{minimize}}

\title{
DeepZero: Scaling Up Zeroth-Order Optimization for Deep Model Training
}

\author{Aochuan Chen$^{\dag, \star}$ ~~Yimeng Zhang$^{\dag, \star}$ ~~Jinghan Jia$^\dag$  ~~James Diffenderfer$^\ddag$ ~~Jiancheng Liu$^\dag$\\ \textbf{Konstantinos Parasyris$^\ddag$} ~~\textbf{Yihua Zhang$^\dag$} ~~\textbf{Zheng Zhang$^\S$} ~~\textbf{Bhavya Kailkhura$^\ddag$} ~~\textbf{Sijia Liu$^\dag$} \\
  $^\dag$Michigan State University,
  $^\ddag$Lawrence Livermore National Laboratory, 
  $^\S$UC Santa Barbara\\
    $^*$Equal contributions
}

\iclrfinalcopy
\begin{document}

\maketitle

\begin{abstract}
Zeroth-order (ZO) optimization has become a popular technique for solving machine learning (ML) problems when first-order (FO) information is difficult or impossible to obtain. However, the scalability of ZO optimization remains an open problem: Its use has primarily been limited to relatively small-scale ML problems, such as sample-wise adversarial attack generation. To our best knowledge, no prior work has demonstrated the effectiveness of ZO optimization in training deep neural networks (DNNs) without a significant decrease in performance. To overcome this roadblock, we develop \textit{\DeepZero}, a principled ZO deep learning (DL) framework that can scale ZO optimization to DNN training from scratch through three primary innovations.
\textit{First}, we demonstrate the advantages of coordinate-wise gradient estimation ({\CGE}) over randomized vector-wise gradient estimation in training accuracy and computational efficiency. 
\textit{Second},  we propose a sparsity-induced ZO training protocol that extends the model pruning methodology using only finite differences to explore and exploit the sparse DL prior in {\CGE}. 
\textit{Third}, we develop the methods of feature reuse and forward parallelization to advance the practical implementations of ZO training.
Our extensive experiments show that DeepZero achieves state-of-the-art (SOTA) accuracy on ResNet-20 trained on CIFAR-10, approaching FO training performance for the first time. Furthermore, we show the practical utility of DeepZero in applications of certified adversarial defense and DL-based partial differential equation error correction, achieving 10-20\% improvement over SOTA. We believe our results will inspire future research on scalable ZO optimization and contribute to advancing  DL with black box. Codes are available at \url{https://github.com/OPTML-Group/DeepZero}.
\end{abstract} 

\vspace*{-3mm}
\section{Introduction}
\vspace*{-3mm}

In the realm of machine learning (ML), optimization algorithms have played a crucial role in enabling the training of complex models, yielding unprecedented insights and predictive capabilities across diverse domains. Over the years, first-order (FO) gradient-based methods, such as stochastic gradient descent (SGD) and its variants \citep{gardner1984learning, amari1993backpropagation, bottou2010large, bottou2012stochastic}, have become the default choice for model training. These methods rely on gradient information to iteratively update model parameters, aiming to minimize a given loss function. Nonetheless, several practical settings exist where FO gradient information is either unavailable or infeasible to compute, calling for alternative strategies.
Zeroth-order (ZO) optimization \citep{flaxman2005online,shamir2013complexity,ghadimi2013stochastic,nesterov2015random,duchi2015optimal,liu2018zeroth,ilyas2018prior, zhang2024revisiting}
has emerged as a promising approach to address these challenges, as it leverages finite differences of function values to estimate gradients, rather than requesting explicit gradient information. 
Therefore, with minor modifications to FO algorithms, ZO optimization can be applied to various real-world circumstances where FO gradients are difficult to obtain. 
For example, in disciplines like physics and chemistry, ML models may interact with intricate simulators or experiments where the underlying systems are non-differentiable \citep{thelen2022comprehensive, tsaknakis2022zeroth, louppe2019adversarial, souza2023exploring, baydin2020differentiable}.  
Additionally, black-box learning scenarios often arise when deep learning (DL) models are integrated with third-party APIs, such as adversarial attack and defense against black-box DL models \citep{chen2017zoo,ilyas2018black,zhang2022robustify,verma2023certified} and black-box prompt learning for language-model-as-a-service \citep{diao2022black,sun2022black}.
Furthermore, the principled backpropagation (BP) mechanism  \citep{amari1993backpropagation, rumelhart1995backpropagation} for calculating FO gradients may also not be supported when implementing DL models on hardware systems \citep{gu2021efficient, tavanaei2019deep, greengard2020neuromorphic, jabri1992weight, gu2021l2ight}.
In addition to ZO optimization, another relevant research direction in the field of DL focuses on developing biologically-plausible, BP-free methods. Examples include forward gradient-based methods \citep{ren2022scaling, baydin2022gradients, silver2021learning, belouze2022optimization}, greedy layer-wise learning \citep{nokland2019training},  and Hebbian learning \citep{isomura2018error, moraitis2022softhebb}. However, these techniques require access to computational graphs and are highly dependent on the used DL software frameworks and/or model architectures. In contrast, ZO optimization solely relies on model queries and is free of computational graphs utilized. As a result, ZO optimization has broad applicability to DL problems that involve black-box query-only components.
Despite the promise of ZO optimization, scalability bottlenecks hinder its application in medium or large-scale DNN training 
\citep{wang2017stochastic, liu2018zeroth, ohta2020sparse, cai2021zeroth, zhang2022robustify}.  
As problem dimensionality increases, the accuracy and efficiency of traditional ZO methods deteriorate. 
This is because ZO finite difference-based gradient estimates are biased estimators of FO gradients, and the bias becomes more pronounced in higher-dimensional spaces \citep{liu2018zeroth, cai2021zeroth,balasubramanian2018zeroth}.

\begin{figure}[t]
    \centering
    \includegraphics[width=0.9\linewidth]{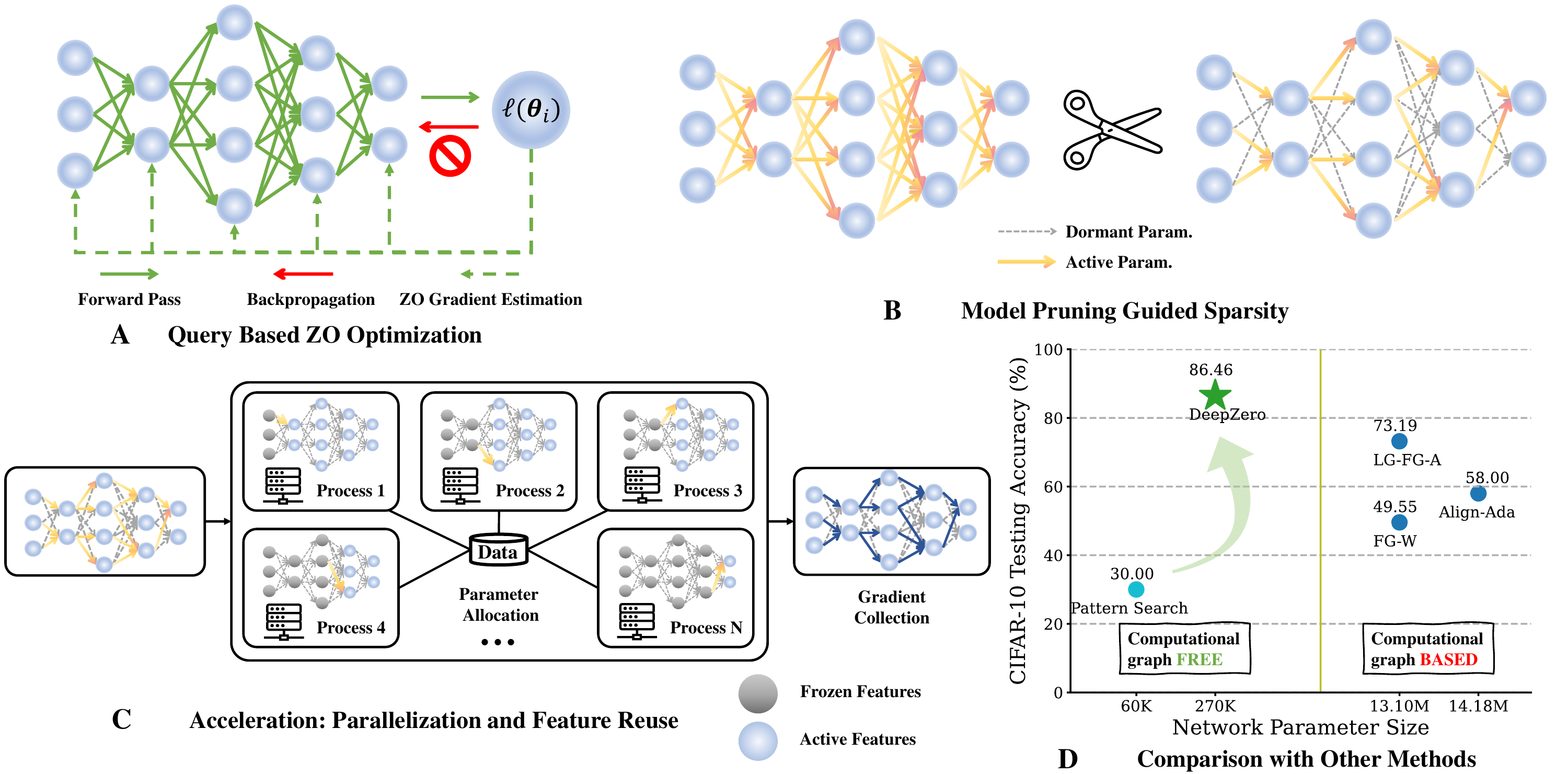}
    \caption{\footnotesize{
    Overview of our {\DeepZero} framework.
    \textbf{A:} ZO gradient estimation via model queries (Sec.\,\ref{sec: ZO_opt_backgroup}).
    \textbf{B:} Model pruning 
    guides gradient sparsity (Sec.\,\ref{sec: sparse_ZO_train}).
    \textbf{C:} Acceleration by parallelization and feature reuse (Sec.\,\ref{sec: acceleration}). 
    \textbf{D:} 
    {\DeepZero} comparison with the computational graph free baseline Pattern Search   \citep{chiang2023loss} and  computational graph dependent methods without BP, Align-Ada \citep{boopathy2022train}, LG-FG-A and FG-W \citep{ren2022scaling}, on  CIFAR-10. 
    }
    }
    \label{fig: overview}
    \vspace*{-8mm}
\end{figure}

These challenges motivate the central question addressed in this work: 
\textit{\textbf{(Q)} How to scale up ZO optimization for training deep models?}
To address \textbf{(Q)}, we propose a novel framework, `{\DeepZero}', which infuses novel model pruning and parallel computing techniques to scale up ZO DNN training (see \textbf{Fig.\,\ref{fig: overview}} for a schematic overview). Our \textbf{main contributions} are summarized below.

\ding{182} We show that deterministic \underline{c}oordinate-wise \underline{g}radient \underline{e}stimation (\CGE) outperforms vector-wise \underline{r}andomized \underline{g}radient \underline{e}stimation (\RGE) in both accuracy and computation efficiency when scaling to deep model training. 
Further, CGE becomes increasingly advantageous as model depth increases.

\ding{183} 
We show that sparsity is crucial for realizing model training via  {\CGE} with finite differences. 
In contrast to prior work, we find that sparsity for black-box models can be obtained for `free' by extending the current pruning-at-initialization technique to the ZO learning paradigm. 
The established synergy between pruning and {\CGE} presents a promising avenue for efficient ZO training of DNNs. 

\ding{184} 
We identify the parallelization-fit property inherent to {\CGE}-based ZO optimization and propose a novel \textit{forward parallelization} method based on this property. Our framework enables \textit{feature reuse} in deep learning, which further accelerates parallel training by eliminating redundant computations.

\ding{185}
We introduce our proposed ZO deep model training framework, `{\DeepZero}'. To demonstrate the empirical superiority of {\DeepZero}, we conduct extensive experiments on both standard image classification benchmarks and real-world black-box DL applications. For example, when employing {\DeepZero} to train a ResNet20 on CIFAR-10, we obtain $86.94\%$ testing accuracy, the best reported in the literature of gradient-free model training. 
We also exemplify the vast potential and practical impact of {\DeepZero} in two real-world DL tasks: black-box defense for adversarial robustness \citep{zhang2022robustify} and physics-informed DL with solver-in-the-loop \citep{um2020solver}. 

To clarify, our work aims to extend the scalability of ZO optimization for DL applications, addressing cases where FO optimization becomes challenging or infeasible.
 Yet, it is essential to note that the proposed advancements in ZO training are not intended to overcome the ultimate scalability challenges to train deep networks at any scale.

\vspace*{-4mm}
\section{Related Work}
\vspace*{-3mm}
 
\textbf{Classical gradient-free optimization.} Early research efforts can be broadly categorized into two groups: direct search-based methods (DSMs) and model-based methods (MBMs) \citep{wright1999numerical,conn2009introduction,rios2013derivative,larson2019derivative}. DSMs include techniques like coordinate \citep{fermi1952numerical} and pattern search \citep{torczon1991convergence} methods and the Nelder-Mead simplex method \citep{nelder1965simplex}. MBMs consist of model-based descent \citep{bortz1998simplex} and trust region \citep{conn2000trust} methods. Evolutionary optimization offers a generic population-based gradient-free computational framework including genetic algorithms \citep{grefenstette1993genetic} and particle swarm optimization \citep{vaz2009pswarm}. 
Bayesian optimization \citep{shahriari2015taking,eriksson2019scalable} has garnered recent attention by using a Gaussian process (GP) to fit a black-box objective function and estimate an optimization solution. However, acquiring an accurate GP is computationally intensive. 
 
\textbf{Zeroth-order optimization.}
In contrast to classical gradient-free methods, ZO optimization approximates gradients using finite differences, simplifying implementation by minimizing modifications of 
FO gradient-based algorithms. 
Like FO methods, ZO enjoys provable convergence guarantees \citep{nesterov2017random,duchi2015optimal,liu2020primer}.
ZO optimization has gained significant attention for its success in solving various emerging ML problems \citep{ghadimi2013stochastic, nesterov2015random, flaxman2005online, duchi2015optimal}. 
Examples include adversarial attack and defense \citep{chen2017zoo, tu2019autozoom,ye2018hessian,ilyas2018black,zhang2022robustify,verma2023certified, zhao2019design, hogan2018universal, shu2022zeroth},
model-agnostic contrastive  explanation   \citep{dhurandhar2019model}, visual prompting for transfer learning \citep{tsai2020transfer}, computational graph unrolling \citep{vicol2023low},   automated ML \citep{gu2021optimizing,wang2022zarts},
policy search in reinforcement learning \citep{vemula2019contrasting},  network resource management \citep{liu2018zeroth}, 
ML-based scientific workflow optimization  \citep{tsaknakis2022zeroth}, and on-chip learning \citep{gu2021efficient}. Despite ZO's successes in solving ML problems, its application has been limited to relatively small scales. 
For instance, ZO optimizers used for generating adversarial attacks, contrastive explanations, and visual prompts only operate in the input parameter space, which has the dimension of a single input example. 
Some acceleration techniques have been developed to improve ZO performance in larger problems, such as using historical information to enhance a ZO gradient estimator \citep{meier2019improving,cheng2021convergence}, and exploiting gradient sparsity to reduce ZO dependence on problem size \citep{wang2017stochastic,cai2022zeroth,cai2021zeroth,balasubramanian2018zeroth,ohta2020sparse,gu2021efficient}.
While gradient sparsity has been used to improve scalability~\citep{bartoldson2023compute}, we propose an advanced strategy that leverages model pruning techniques to identify and exploit sparsity in neural network parameters effectively. 
Our approach is less restrictive than traditional gradient sparsity assumptions and allows for greater flexibility in selecting what to prune. 
To the best of our knowledge, no prior work has demonstrated the practicality of scalable ZO optimization for deep model training without significant performance loss compared to the FO counterpart.

\textbf{DL without backpropagation.}
Forward gradient learning \citep{baydin2022gradients,ren2022scaling,silver2021learning,belouze2022optimization}, which builds upon the forward-mode automatic differentiation (AD) capabilities of current DL software frameworks, does not rely on finite differences to approximate FO gradients like ZO optimization. Instead, it relies on forward-mode AD to calculate a forward (directional) gradient. 
This gradient is obtained by projecting the FO gradient onto a direction vector and is an unbiased estimator of the FO gradient \citep{baydin2022gradients}. 
In contrast, ZO gradient estimation based on finite differences is biased \citep{duchi2015optimal,liu2020primer}.
However, one main limitation of forward gradient learning is that it requires full access to AD software and the deep model, making it impractical for solving black-box DL problems. 
Recent advances in \citep{ren2022scaling} further improved the scalability of forward gradient learning by using finer-level model information to design architecture-specific local objective functions.
Other BP-free DL methods are motivated by seeking a biological interpretation of DL but share similar limitations with forward gradient learning. 
Some examples include greedy layer-wise learning \citep{nokland2019training}, input-weight alignment for wide neural networks in the neural tangent kernel (NTK) regime \citep{boopathy2022train}, the Forward-Forward algorithm \citep{hinton2022forward}, Hebbian Learning \citep{isomura2018error, moraitis2022softhebb}, and synthetic gradients \citep{jaderberg2017decoupled}.
\vspace*{-3mm}
\section{ZO Optimization through Function Value-based Gradient Estimation: Randomized or Coordinate-wise?}
\label{sec: ZO_opt_backgroup}
\vspace*{-2mm}

We now introduce the ZO optimization setup 
and discuss two 
ZO gradient estimation schemes:  deterministic coordinate-wise gradient estimation (\textbf{\CGE}) and randomized vector-wise gradient estimation (\textbf{\RGE}). We will demonstrate the advantage of {\CGE} over {\RGE} for DNN training. 
This inspires further improvements 
for scaling {\CGE}-based ZO optimization.

\textbf{ZO optimization and gradient estimation.}
Let $\ell(\boldsymbol \theta)$ denote a \textbf{loss function} that we want to minimize over the \textbf{optimization variables} $\btheta \in \mathbb R^d$ (\textit{e.g.}, model parameters of a neural network).
The ZO   optimizer interacts with the objective function $\ell$ only by submitting inputs (\textit{i.e.}, realizations of $\boldsymbol \theta$) and receiving the corresponding function values. It slightly modifies the commonly-used first-order (FO) gradient-based algorithm by approximating the FO gradient through function value-based gradient estimates \citep{liu2020primer}. 
This 
is essential when explicit differentiation is difficult due to the black-box nature of the loss function 
\citep{zhang2022robustify,liu2020min,chen2017zoo}, or
when explicit differentiation is undesired due to concerns about energy efficiency \citep{gu2021efficient,liu2018zeroth_admm}.
{\RGE} \citep{nesterov2017random,ghadimi2013stochastic,duchi2015optimal,spall1992multivariate} and {\CGE} \citep{kiefer1952stochastic,lian2016comprehensive,berahas2022theoretical} are two commonly-used  gradient estimators based on finite differences of $\ell$. 
{\RGE} acquires finite differences via random perturbations of 
$\boldsymbol \theta$
while {\CGE} uses deterministic coordinate-wise perturbations of $\boldsymbol \theta$ \citep{liu2020primer}. Their formal definitions are given by 

\vspace*{-5mm}
{\small{
\begin{align}
    \text{(\textbf{\RGE})} ~~ \hat{\nabla}_{\btheta} \ell(\btheta) = \frac{1}{q} \sum_{i=1}^q \left [ \frac{\ell(\btheta + \mu \mathbf u_i) - \ell(\btheta)}{\mu}  \mathbf u_i \right ];  ~
    \text{(\textbf{\CGE})} ~~ \hat{\nabla}_{\btheta} \ell(\btheta) = \sum_{i=1}^d \left [ \frac{\ell(\btheta + \mu \mathbf e_i) - \ell(\btheta)}{\mu}  \mathbf e_i \right ],
    \label{eq: RGE_CGE}
\end{align}}}%
\begin{wrapfigure}{r}{50mm}
\vspace*{-5mm}
\centerline{
\includegraphics[width=50mm,height=!]{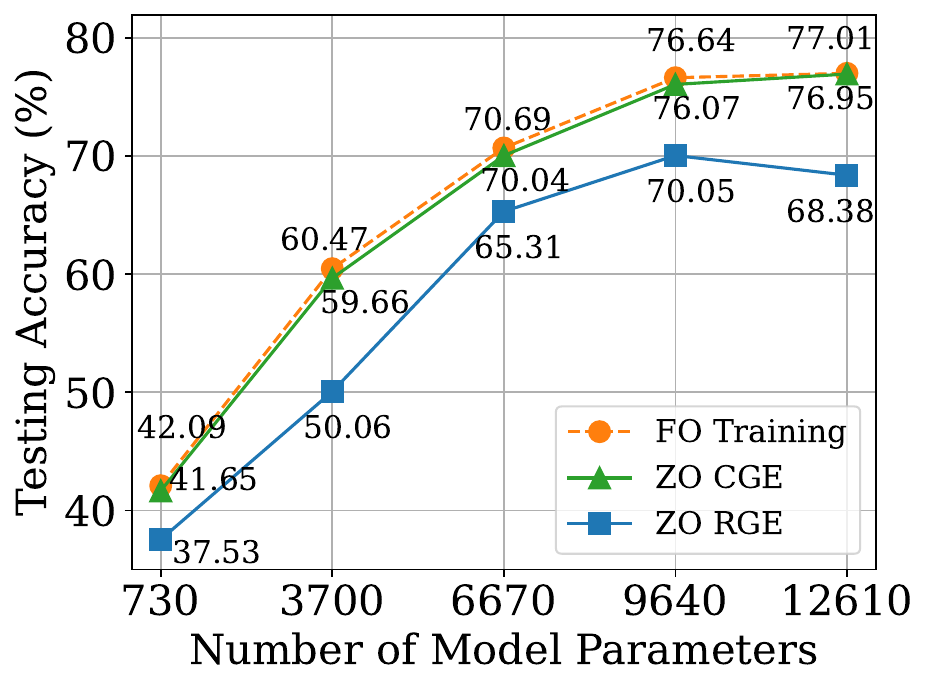}
}
\vspace*{-3mm}
\caption{\footnotesize{ 
Performance comparison of training a simple CNN with varying numbers of parameters on   CIFAR-10   using different training methods.
}}
\label{fig: Acc_CGE_RGE}
\vspace*{-6mm}
\end{wrapfigure}%
where  $\hat{\nabla}_{\btheta} \ell$ denotes an estimation of the FO gradient $\nabla_{\btheta}\ell$ with respect to $\btheta$. In {({\RGE})}, $\mathbf u_i $ denotes a randomized perturbation vector, \textit{e.g.},  drawn from the standard Gaussian distribution $\mathcal N(\mathbf 0, \mathbf I)$, $\mu > 0$ is a perturbation size (\textit{a.k.a.} smoothing parameter), and $q$ is the number of random directions used to acquire finite differences. 
In {(\CGE)}, $\mathbf e_i$ denotes a standard basis vector, and $\frac{\ell(\btheta + \mu \mathbf e_i) - \ell(\btheta)}{\mu}$ provides the finite-difference estimation of the partial derivative of $\ell(\btheta)$ at the $i$th coordinate $\btheta_i$.
Finite difference approximations in \eqref{eq: RGE_CGE} are motivated by \textit{directional derivatives}. Take {\RGE} (with $q = 1$) as an example. As $\mu \to 0$, finite difference in {\RGE}  converges to directional derivative $\ell^\prime(\btheta) \Def \mathbf u^T \nabla_{\btheta} \ell(\btheta) = \lim_{\mu \to 0} \frac{\ell(\btheta + \mu \mathbf u_i) - \ell(\btheta)}{\mu} $ of the function $\ell$ at the point $\btheta$ in the direction $\mathbf u$ \citep{urruty1993convex}. As a result, the expression 
$ \ell^\prime(\btheta) \mathbf u$ yields 
$
\mathbb E[    \ell^\prime(\btheta) \mathbf u ] = \mathbb E[  (\mathbf u \mathbf u^T) \nabla_{\btheta} \ell(\btheta)  ]  = \nabla_{\btheta} \ell(\btheta)
$ (recall that $\mathbb E[ \mathbf u \mathbf u^T] = \mathbf I$). 
This implies that $\ell^\prime(\btheta) \mathbf u$ is an \textit{unbiased} gradient estimator of $\nabla_{\btheta} \ell(\btheta)$ and its \textit{biased} finite difference approximation is given by \eqref{eq: RGE_CGE} \citep{duchi2015optimal}.

\textbf{{\RGE} or {\CGE}?}  
First, the \textit{function query costs} for {\RGE} and {\CGE} differ, with {\RGE} taking $O(q)$ queries and {\CGE} taking $O(d)$ queries based on \eqref{eq: RGE_CGE}.
Compared to {\CGE}, {\RGE} has the flexibility to specify $q < d$ to reduce the number of function evaluations. 
Despite the query efficiency, it remains uncertain whether {\RGE} can deliver satisfactory accuracy when training a deep 
model from scratch.
To this end, we undertake a preliminary investigation wherein we train a basic convolutional neural network (CNN) of different sizes on CIFAR-10, employing both {\RGE} and {\CGE}. To ensure a fair comparison in query complexity, we set the query number $q$ in {\RGE} equal to the problem size $d$ used in {\CGE}.
\textbf{Fig.\,\ref{fig: Acc_CGE_RGE}} presents the test accuracy of the learned CNN against the number of model parameters (equivalent to the number of model queries). 
Here the training recipe 
is specified by the FO SGD, the ZO {\RGE}-based SGD, and the ZO {\CGE}-based SGD. 
We observe that {\CGE} can achieve test accuracy comparable to FO training and significantly outperforms {\RGE}.  
This experiment highlights the superiority of {\CGE} over {\RGE} in terms of optimization accuracy even when the latter uses $q = d$. This \textit{accuracy merit} of {\CGE} is particularly valuable when training  more complex neural networks.
In Appx.\,\ref{app: zo}, we provide a detailed analysis of the computational costs using {\CGE} vs. {\RGE}. The time cost relative to model depth is shown in Fig.\,\ref{fig: Time_CGE_RGE}. And Tab.\,\ref{tab: rge_cge_time_compare} assesses gradient estimation time costs. We find that {\CGE} demonstrates greater time efficiency than RGE. The computation efficiency loss of {\RGE} is in that it needs to generate and integrate a $d$-dimension perturbation vector into the entire model at once every query.
Based on the advantages of {\CGE} over {\RGE} in terms of both accuracy and computation efficiency, we choose {\CGE} as the preferred ZO gradient estimator. 
However, query complexity of {\CGE} is still a bottleneck, as it scales with model size $d$.

\vspace*{-4mm}
\section{Sparsity-Assisted ZO Training: A Pruning Lens and Beyond}
\label{sec: sparse_ZO_train}
\vspace*{-3mm}
One valuable property of {\CGE} is the disentanglement of finite differences across coordinates, which suggests that reducing {\CGE}'s query complexity is aligned with pruning the model weights that are being optimized. 
With this in mind, we propose integrating ZO optimization with {\textit{pruned gradients}} to design a more effective inductive bias for ZO deep model training.
It is worth noting that the sparsity has been explored in several existing ZO optimization methods to improve the query efficiency of gradient estimation \citep{wang2017stochastic,cai2022zeroth,cai2021zeroth,balasubramanian2018zeroth,ohta2020sparse,gu2021efficient}. 
However, prior work suffered from two main limitations. Firstly, exact sparsity was assumed in the original FO gradients, which required an additional sparse learning method (such as LASSO \citep{wang2017stochastic}) to recover these sparse gradients from function queries. 
Secondly, it remains unclear how to optimize the sparsity pattern via a ZO oracle, as the existing method calls for overly heuristics-based pruning methods (e.g., random \citep{gu2021efficient} or magnitude \citep{ohta2020sparse, zhang2022data} pruning). 
Overly increasing sparsity ultimately limits optimization performance. 
In what follows, we propose a new pruning approach that relies only on model queries, enjoys computation efficiency, and can improve ZO optimization accuracy by inducing an appropriate gradient sparsity.

\textbf{{\zograsp}: Model pruning via ZO oracle.} 
The compressibility of model weights for DL has been extensively studied \citep{han2015deep, frankle2018lottery,  
ma2021sanity, 
zhang2022advancing, zhang2022data,blalock2020state,tanaka2020pruning,lee2018snip,wang2020picking,su2020sanity, diffenderfer2021winning}. For instance, the lottery ticket hypothesis \citep{frankle2018lottery} demonstrated that a randomly initialized, dense neural network contains a high-quality \textit{sparse subnetwork}. 
However, current effective pruning methods incorporate model training as an intermediate step \citep{frankle2018lottery,  ma2021sanity, zhang2022advancing, diffenderfer2021multi}. \textit{Thus, they are not well-suited for finding sparsity via a ZO oracle.}

To address the above challenge, we draw inspiration from training-free pruning methods, known as pruning-at-initialization \citep{tanaka2020pruning,lee2018snip,wang2020picking}. Within this family, gradient signal preservation (\grasp) \citep{wang2020picking} is a method to identify the sparsity prior of DL through the gradient flows of a randomly-initialized network. 
While {\grasp} still requires the FO and second-order derivative information, we can estimate these derivatives using {only function queries} to design the ZO version of {\grasp} (termed  \textbf{\zograsp}). Specifically, {\grasp} \citep{wang2020picking} assigns pruning scores (denoted by $\mathbf S$) to  model initialization $\btheta$. These scores reflect the change in gradient flow after pruning the weights:

\vspace*{-5mm}
{\small{
\begin{align}
\mathbf S = - \boldsymbol{\theta} \odot (\mathbf H \mathbf g), ~~ \mathbf H = \nabla_{\boldsymbol{\theta}, \boldsymbol{\theta}}^2 \ell (\boldsymbol{\theta}) 
, ~\mathbf g = \nabla_{\boldsymbol{\theta}} \ell (\boldsymbol{\theta}) 
,
\label{eq: grasp}
\end{align}
}}%
where recall that $\ell$ is the loss function of model training, $\odot$ denotes the entry-wise multiplication, and $\mathbf H \mathbf g$ represents the Hessian-gradient product.
Using the ZO learning paradigm, we can first approximate the Hessian-gradient product as the finite difference between two gradients (\textit{i.e.}, $\nabla_{\boldsymbol{\theta}}\ell( \boldsymbol{\theta} + \mu \mathbf g  )$ and $\nabla_{\boldsymbol{\theta}} \ell(\boldsymbol{\theta})$), in the direction $\mathbf g$ with the smoothing parameter $\mu$.  
Second, we replace the FO gradient $\nabla_{\btheta} \ell$ with the ZO gradient estimate $\hat{\nabla}_{\btheta} \ell$ given in \eqref{eq: RGE_CGE}. Combining this yields  {\zograsp}:

\vspace*{-5mm}
{\small { \begin{align}
\hat{\mathbf S} \Def  -\btheta \odot 
 \frac{ \hat{\nabla}_{\boldsymbol{\theta}}\ell( \boldsymbol{\theta} + \mu \hat{\mathbf g}  ) - \hat{\nabla}_{\boldsymbol{\theta}} \ell(\boldsymbol{\theta}) }{\mu}.
    \label{eq: grad_vec_product_approx1}
\end{align}}}
\vspace*{-5mm}%

In practice, we found that the pruning mask determined by ranking the entries in $\hat{\mathbf S}$ is resilient to the ZO gradient estimation error. 
Therefore, we utilize {\RGE} with a relatively small number of queries ($q < d$) to implement {\zograsp}. 
This reduces the function query cost without compromising pruning performance; see \textbf{Tab.\,\ref{tab: pruning_com_rn20}} and \textbf{Tab.\,\ref{tab: pruning_com_rn18}} for empirical justifications.  
Our results show that {\zograsp} significantly outperforms random pruning and yields pruned models with accuracy comparable to FO-{\grasp}.

\textbf{Integrating sparsity with {\CGE}.}
As finite differences in {\CGE} \eqref{eq: RGE_CGE} are decomposable over weights, it is easy to incorporate sparsity into {\CGE}. 
To retain the accuracy benefits of training dense models, we incorporate gradient sparsity (in {\CGE}) rather than weight sparsity. This ensures that we train a dense model in the weight space, rather than training a sparse model where the sparsity determined by {\zograsp} is directly applied.
Let $\mathcal S_{\zograsp}$ be the coordinate set of unpruned model weights found by   {\zograsp}. 
The sparsity-induced {\CGE} is given by

\vspace*{-5mm}
{\small{
 \begin{align}
     \hat \nabla_{\boldsymbol \theta}   \ell ({\boldsymbol \theta} ) =  \sum_{i \in \mathcal S_{\zograsp}
     } \left [ \frac{\ell ({\boldsymbol \theta}   + \mu \mathbf e_i ) - \ell({\boldsymbol \theta} )}{\mu} \mathbf e_i \right ].
     \tag{\text{Sparse-{\CGE}}}
     \label{eq: CGE_sparsity}
 \end{align}
 }}%
 It is clear that  \eqref{eq: CGE_sparsity} reduces the query complexity of the original {\CGE} from $O(d)$ to $O(|\mathcal S_{\zograsp}|)$, where $|\mathcal S_{\zograsp}|$ denotes the cardinality of the coordinate set $\mathcal S_{\zograsp}$. 
There may exist two direct methods
for integrating \eqref{eq: CGE_sparsity} into ZO optimization. 
$\mathcal M_1$: This method involves \textit{alternating} between {\zograsp} and {\CGE}-based ZO optimization. 
At each iteration, $\mathcal S_{\zograsp}$ is updated based on the model weights from the previous iteration and then used to construct \eqref{eq: CGE_sparsity} for updating $\btheta$ at the current iteration. 
$\mathcal M_2$: This method involves performing \textit{pruning before ZO training}. 
That is, {\zograsp} is conducted at model initialization, and the resulting $\mathcal S_{\zograsp}$ is applied to \eqref{eq: CGE_sparsity} and kept fixed during training.
\textbf{Both $\mathcal M_1$ and $\mathcal M_2$ have limitations.} 
$\mathcal M_1$ requires repeated calls to {\zograsp} to update $\mathcal S_{\zograsp}$, leading to a higher query cost for ZO model training. 
$\mathcal M_2$ addresses the query complexity by performing {\zograsp} before training, but it can only produce a smaller model after training. It is known that heavily-pruned models suffers from performance degradation (\textit{e.g.}, 95\% sparse model in 
Tab.\,\ref{tab: pruning_com_rn20} in 
Appx.\,\ref{app: pruning}).
Thus, it is nontrivial to
integrate {\zograsp} with ZO training due to the requirement of balancing query efficiency and training effectiveness. To address this, we propose 
\textbf{{\zograsp}-oriented dynamic sparsity pattern}, which leverages  {\zograsp}  to determine layer-wise pruning ratios (LPRs) that can capture DNN compressibility. 
This approach shares a similar essence with smart ratio introduced in \citep{su2020sanity}.
Specifically, 
we acquire LPRs from {\zograsp} at randomly initialized weights prior to ZO training, which is query-efficient like $\mathcal M_2$. 
However, unlike $\mathcal M_2$, LPRs allow for random shuffling of sparse gradient positions in $\btheta$ only if these LPRs are obeyed. 
This allows us to mimic $\mathcal M_1$ to alternate between model weight updates and $\mathcal S_{\zograsp}$ updates, with the latter achieved by LPR-guided randomly updated sparsity patterns. 
Thus, ZO optimization can train the dense model using iteratively-updated {\eqref{eq: CGE_sparsity}} with   LPRs-guided dynamic sparsity patterns. 
Overall, our proposal has the query efficiency of $\mathcal M_2$ with the training effectiveness of $\mathcal M_1$, resulting in a balanced integration of {\zograsp} into ZO training. 
We summarize the algorithmic pipeline in \textbf{Algorithm\,\ref{alg: lpr_zo}} in Appx.\,\ref{app: alg_details}, where CGE and {\zograsp}-oriented dynamic sparsity pattern are clearly described in a unified framework. We also refer readers to Appx.\,\ref{app: alg_details} for more explanation and comparisons with $\mathcal M_1$ and $\mathcal M_2$.
We provide a convergence rate analysis in Appx.\,\ref{app: convergence_rate}.

\vspace*{-2mm}
\section{Improving Scalability: Feature Reuse \& Forward Parallel}
\label{sec: acceleration}
\vspace*{-2mm}

\begin{wrapfigure}{r}{.3\textwidth}
\vspace*{-3mm}
\centerline{
\includegraphics[width=0.3\textwidth,height=!]{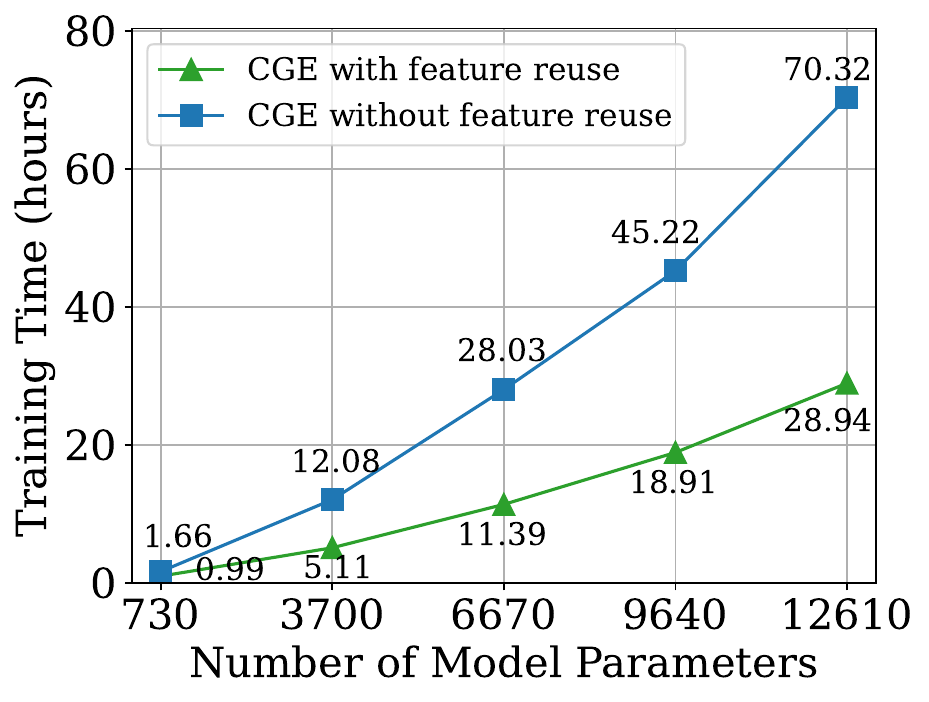}
}
\vspace*{-4mm}
\caption{\footnotesize{Computation cost   of {\CGE}-based ZO training w/ feature reuse vs. w/o feature reuse. The  setup follows Fig.\,\ref{fig: Time_CGE_RGE}.}}
\label{fig: feature_reuse}
\vspace*{-3mm}
\end{wrapfigure}%

We investigate two characteristics of ZO training that can further enhance implementation scalability: \textit{feature reuse} and \textit{forward parallelization}. 
The former disentangles intermediate features from weight perturbations, while the latter uses the finite-difference nature in {\CGE} to enable scalable distributed implementation. 

\textbf{Reusing intermediate features.} 
As shown in \eqref{eq: RGE_CGE}, {\CGE} perturbs each parameter element-wise. 
Thus, 
one can reuse the feature immediately preceding the perturbed layer and carry out the remaining forward pass operations instead of starting from the input layer, as illustrated in \textbf{Fig.\,\ref{fig: overview}}. 
 The above characteristic of {\CGE}-based model training  is referred to as `\textbf{feature reuse}'.
 More concretely, let $f_{\btheta}(\mathbf x)$ be  a deep model with parameters $\btheta$   and   input $\mathbf x$. We can express $f_{\btheta}(\mathbf x)$ as a multi-layer composite function  

\vspace*{-5mm}
{\small{
\begin{align}
 \hspace*{-2mm}   f_{\btheta}(\mathbf x) = f_{\btheta_{>l}}(\mathbf z_l) = \underbrace{f_{\btheta_L} \circ f_{\btheta_{L-1}} \circ \cdots \circ f_{\btheta_{l+1}}}_\text{$f_{\btheta > l}(\cdot)$}  \circ \underbrace{ f_{\btheta_l} \circ \cdots \circ f_{\btheta_1}(\mathbf x)}_\text{$\mathbf z_l = f_{\btheta_{1:l}}(\mathbf x)$},
  \hspace*{-2mm}     \label{eq: f_network_layers}
\end{align}
}}%
where $f_{\btheta_l}$ denotes the model's $l$th layer, $L$ is the total number of model layers,  and 
$\circ$ is the function composition operation.
Based on \eqref{eq: f_network_layers}, 
if coordinate-wise weight perturbations are applied to the $(l+1)$th layer  and its subsequent layers (\textit{i.e.}, $\btheta_{> l}$), 
the model's outputs corresponding to these perturbed weights can be efficiently obtained by keeping the intermediate features up to the $l$th layer (\textit{i.e.}, $\mathbf z_l$) intact. This efficiency becomes more pronounced when perturbing the parameters of deeper 
layers (\textit{i.e.}, for a larger $l$). \textbf{Fig.\,\ref{fig: feature_reuse}} compares the runtime of {\CGE}-based ZO training with and without feature reuse. 
Empirically, CGE with feature reuse exhibits a $2\times$ reduction in training time.

\textbf{Parallelization of coordinate-wise finite differences.} 
{\CGE} enables parallelization of model training due to its alignment of parameter perturbations with forward passes.
If there exist $M$ processes (across multiple GPUs), we can decompose {\CGE} \eqref{eq: RGE_CGE} based on its parameter coordinates 
yielding

\vspace*{-5mm}
{\small{
\begin{align}
 \hat{\nabla}_{\btheta} \ell(\btheta) = \sum_{i=1}^M  \hat{\mathbf g}_i, ~~ \hat{\mathbf g}_i \Def  
 \sum_{j \in \mathcal S_i } \left [ \frac{\ell(\btheta + \mu \mathbf e_j) - \ell(\btheta)}{\mu}  \mathbf e_j \right ],
    \label{eq: Parallel_CGE}
\end{align}}}%
where $\mathcal S_i$ is the  set of active parameters assigned to process $1 \leq i \leq M$. 
Hence, each process can take $|\mathcal S_i|$ forward passes.
This decoupling property enables scaling forward passes via distributed machines, which can significantly improve training speed. 
We refer to this parallelization for finite differences as `\textbf{forward parallelization}'. 
It is worth noting that forward parallelization is different from the conventional data parallelization used for FO distributed training \citep{goyal2017accurate,you2018imagenet}. 
Our method avoids any performance loss that may result from data parallelization using an overly large batch size, which can cause the optimizer to get stuck in suboptimal local minima due to a lack of stochasticity.

\vspace*{-3mm}
\section{Experiments}
\label{sec: experiments}
\vspace*{-3mm}
In this section,
we first train \textit{ResNet-20} for standard image classification on \textit{CIFAR-10}, demonstrating scalability and generalization capability over existing gradient-free learning methods. 
Second, we apply DeepZero to enhance the robustness of a \textit{black-box} DNN against adversarial attacks, where limited access to the  model is available on the defender's end. 
Lastly, we leverage DeepZero to design a \textit{physics-informed ML} system by incorporating a scientific PDE solver into the training loop for reducing numerical errors, highlighting its capability to address complex scientific problems.

\vspace*{-4mm}
\subsection{Image classification task}
\vspace*{-3mm}

\begin{wrapfigure}{r}{52mm}
\vspace*{-5mm}
\centerline{
\includegraphics[width=52mm,height=!]{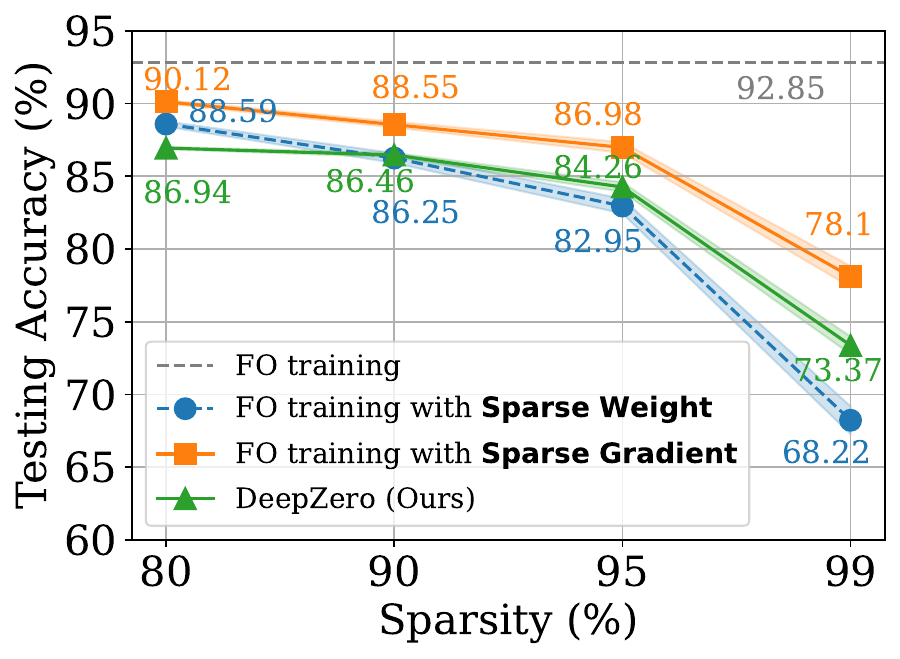}
}
\vspace*{-3.3mm}
\caption{\footnotesize{
Comparison between {\DeepZero} and FO training baselines on a ResNet-20 for CIFAR-10. We report the mean and standard deviation of 3 independent runs for each experiment.
}}
\label{fig: sg_sw}
\vspace{-3mm}
\end{wrapfigure}
\textbf{Experiment setup.}
This study focuses on training ResNet-20 (with 270K parameters) on CIFAR-10 for image classification. We adopt SGD (stochastic gradient descent) as the FO training recipe, with a weight decay of $5\times10^{-4}$ and a momentum of $0.9$. The learning rate is $0.1$, governed by a cosine decay scheduler. In the ZO training scenario,  we replace the FO gradient by \eqref{eq: CGE_sparsity} with a smoothing parameter $\mu = 5 \times 10^{-3}$. When implementing ZO-GraSP \eqref{eq: grad_vec_product_approx1},  we set the query budget $q = 192$ and use the same $\mu$ as {\CGE}. 
Unless specified otherwise, the weight sparsity ratio is chosen to be 90\% and the specific sparsity patterns are determined by {\SR} (Smart Ratio). 
When implementing DeepZero (\textbf{Algorithm\,\ref{alg: deepzo}}), we choose the number of epochs $T = 50$. Experiments are run on 4 NVIDIA V100 GPUs if not specified otherwise. We compare DeepZero with FO training and two SOTA BP-free training: Pattern Search \citep{chiang2023loss} and Input-Weight Alignment (Align-ada) \citep{boopathy2022train}.

\begin{wrapfigure}{r}{59mm}
\vspace*{-5mm}
\centerline{
\includegraphics[width=53mm,height=!]{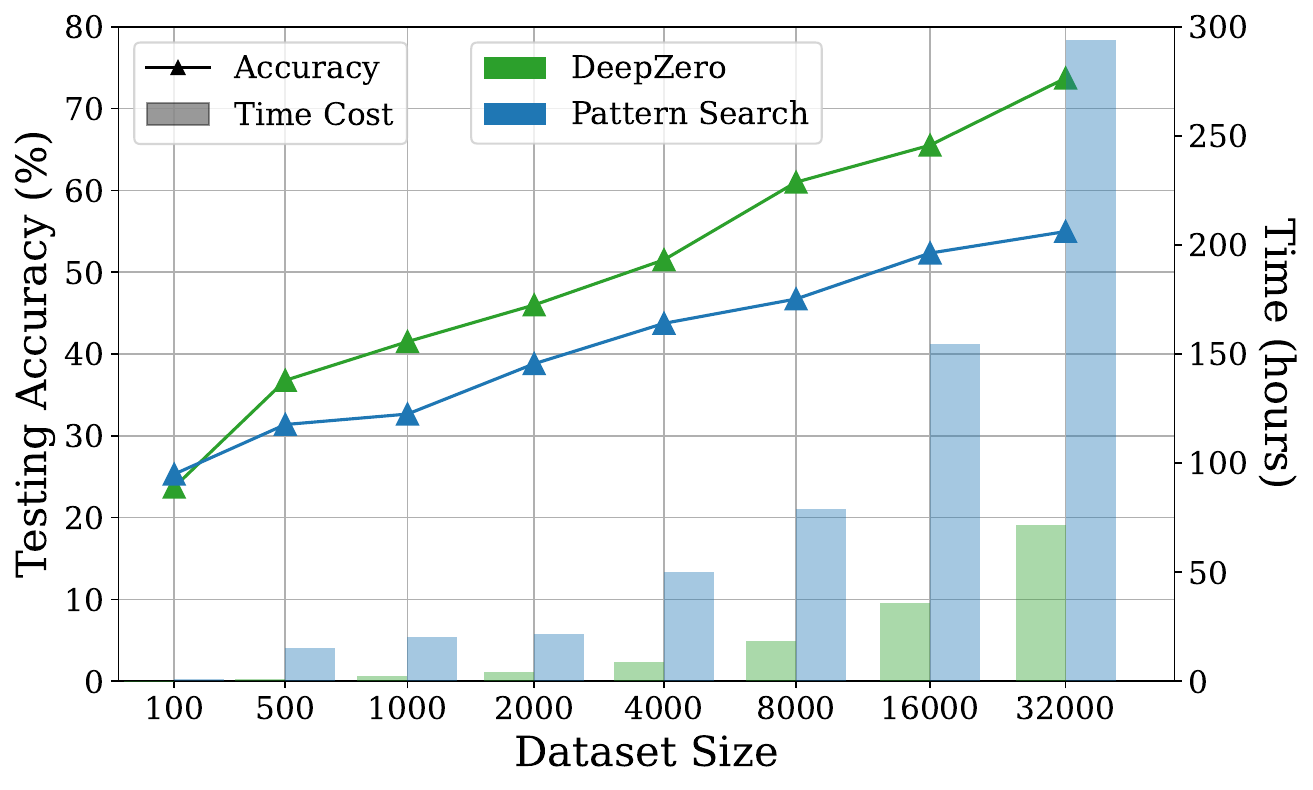}
}
\vspace{-3mm}
\caption{\footnotesize{Comparison of DeepZero and Pattern Search on ResNet-20 for CIFAR-10 with varying dataset sizes. 
All experiments are done on a single NVIDIA A6000 GPU.
}}
  \label{fig: ps}
\vspace{-5mm}
\end{wrapfigure}
\textbf{Comparison with FO training.}
In \textbf{Fig.\,\ref{fig: sg_sw}}, 
we compare the accuracy of DeepZero-trained ResNet-20 with two variants trained by FO recipes: (1) a dense ResNet-20 acquired through FO training and (2) a sparse ResNet-20 acquired through FO training under FO-GraSP sparsity pattern.
As we can see, the accuracy gap still exists between (1) and the model trained with DeepZero in the sparsity regime of 80\% to 99\%. 
This highlights the challenge of ZO optimization for deep model training, where achieving high sparsity is desired to reduce the number of model queries in   \eqref{eq: CGE_sparsity} for scaling to ResNet-20.
Notably, in the sparsity regime of 90\% to 99\%, DeepZero outperforms (2), showcasing the superiority of gradient sparsity in DeepZero compared to weight sparsity (\textit{i.e.}, directly training a sparse model).
In Appx.\,\ref{app: res20_exp}, we provide the DeepZero training trajectory (\textbf{Fig.\,\ref{fig: training_traj}}), performance vs.   data batch   setup (\textbf{Fig.\,\ref{fig: batch_size}}) and training time vs. GPU count (\textbf{Fig.\,\ref{fig: gpu_number}}).

\textbf{Comparison with pattern search \citep{chiang2023loss}.}
In \textbf{Fig.\,\ref{fig: ps}}, we compare the accuracy and runtime cost of DeepZero with Pattern Search \citep{chiang2023loss} for deep model training. 
Pattern
Search has been shown to achieve
comparable test accuracy to SGD in low 
sample regimes, however, effectiveness as the number of data samples increases remains unclear. 
To investigate this, we 
conducted experiments using DeepZero (with 90\% gradient sparsity) and pattern search on ResNet-20 with CIFAR-10, with varying dataset sizes from 100 to 32K. 
We maintained a fixed total epoch number of 40 for both methods to ensure a fair comparison. 
The results demonstrate DeepZero outperforms Pattern Search in all data regimes, except in the case of 100. 
Further, the improvement of DeepZero over pattern search (in both accuracy and efficiency) expands dramatically with increasing dataset sizes, indicating the superior scalability of DeepZero.

\begin{wraptable}{r}{65mm}
\vspace*{-4mm}
\centering
\caption{\footnotesize{Performance of DeepZero vs. BP-free methods on a 8-layer CNN w/ different widths \citep{boopathy2022train}.} }
\vspace*{-2mm}
\label{tab: ntk_cnn}
\resizebox{65mm}{!}{%
\begin{tabular}{c|c|c|c|c|c|c|c|c}
\toprule[1pt]
\midrule
Method & \multicolumn{2}{c|}{DeepZero} & \multicolumn{2}{c|}{FA} & \multicolumn{2}{c|}{DFA} & \multicolumn{2}{c}{Align-ada}  \\ 
\midrule
Width  & 32 & 64 & 64 & 512 & 64 & 512 & 64 & 512 \\
\midrule
Accuracy & 57.7 & \textbf{64.1} & 46.5 & 45.4 & 49.9 & 54.1 & 49.9 & 58.0  \\
\midrule
Time (h) & 4.34 & 28.15 & 0.36 & 3.79 & 0.42 & 3.52 & 0.48 & 4.59  \\
\midrule
\bottomrule[1pt]
\end{tabular}%
}
\vspace*{-4mm}
\end{wraptable}
\textbf{Comparison with input-weight alignment \citep{boopathy2022train}.}
In \textbf{Tab.\,\ref{tab: ntk_cnn}}, we present a comparison between DeepZero 
and the   \textit{Align-ada} approach
\citep{boopathy2022train} for training   neural networks without BP on CIFAR-10.
While other BP-free training methods \citep{lillicrap2014random, nokland2016direct, baydin2022gradients} exist, Align-ada 
stands out as it applies to training \textit{wide} neural networks
and achieves state-of-the-art performance on CIFAR-10, \textit{e.g.}, surpassing 
methods such as feedback alignment (FA) \citep{lillicrap2014random}  and direct feedback alignment (DFA) \citep{nokland2016direct}. 
To ensure fairness, we apply DeepZero to the 8-layer CNN architecture from \citep{boopathy2022train} and compare performance with Align-ada 
at varying model widths. 
We note that the 512-width network was the widest model trained using Align-ada. 
In contrast, the largest width network we train with DeepZero is 64. 
Our results clearly show that DeepZero achieves significantly higher testing accuracy compared to Align-ada, even when training with narrower networks. 
This demonstrates that the improved performance of DeepZero is attributed to its inherent optimization advantages, rather than relying solely on the use of wider networks. 
Lastly, it is worth noting that Align-ada and other BP-free methods rely on access to computational graphs, making them efficient but unsuitable for black-box applications. 

\vspace*{-3mm}
\subsection{Other black-box applications}
\vspace*{-3mm}

\begin{wraptable}{r}{53mm} 
\centering
\vspace*{-4.5mm}
\caption{\footnotesize{CA (\%)  vs.  $\ell_2$-norm based perturbation radius  on ImageNet-10 using FO DS-based defense (FO-DS) \citep{salman2020denoised}, ZO-AE-DS \citep{zhang2022robustify}, and our proposed DeepZero. 
}}
\label{tab: zo_ds_r}
\vspace*{-1mm}
\resizebox{53mm}{!}{%
\begin{tabular}{c|c|c|c}
\toprule[1pt]
\midrule
\multicolumn{4}{c}{ImageNet (10 classes)}  \\
\midrule
Radius $r$ & FO-DS & ZO-AE-DS & DeepZero\\
 \midrule
0.0   & 89.33 & 63.60 & 86.02   \\
0.25  & 81.67 & 52.80 & 76.61  \\
0.5 & 68.87 & 43.13 & 61.80    \\
0.75  & 49.80 & 32.73 & 43.05  \\
 \midrule
\bottomrule[1pt]
\end{tabular}
}
\vspace*{-5mm}
\end{wraptable}

\textbf{Black-box defense against adversarial attacks.}
The black-box defense problem arises when the owner of an ML model is unwilling to share the model details with the defender against adversarial attacks \citep{zhang2022robustify,verma2023certified}. 
This poses a challenge for existing robustness enhancement algorithms \citep{madry2017towards,cohen2019certified,salman2020denoised} that directly robustify {white-box} ML models using FO training. 
To overcome this challenge, ZO optimization was introduced in \citep{zhang2022robustify} to design a \textit{white-box} defense operation given a \textit{query-based black-box} image classifier. 
To address dimensionality challenges with ZO, ZO-AE-DS \citep{zhang2022robustify} introduces an autoencoder (AE) between the \textit{white-box} denoised smoothing (DS) defense operation (to be learned) and the {black-box} image classifier. 
By merging AE's encoder with the black-box module, the dimension of ZO optimization is reduced;
see \textbf{Fig.\,\ref{fig: ae_ds}} in Appx.\,\ref{app: black_box_defense}
for a schematic overview and   \citep{zhang2022robustify} for details. 
The downside of  ZO-AE-DS is  poor scaling to high-resolution datasets (\textit{e.g.}, ImageNet) due to the use of AE, which compromises the fidelity of the image input to the black-box image classifier and leads to inferior defense performance. 
In contrast, DeepZero can directly learn the defense operation integrated with the black-box classifier, without needing AE. 
\textbf{Tab.\,\ref{tab: zo_ds_r}} compares the defense performance of DeepZero with FO defense (DS \citep{salman2020denoised}) and ZO-AE-DS \citep{zhang2022robustify}. 
To ensure a fair comparison, we used the same number of queries (1152) per gradient estimation. 
Following \citep{zhang2022robustify}, we selected a 10-class subset of ImageNet as the training set. The AE and black-box classifier are DnCNN \citep{zhang2017beyond} and ResNet-50, respectively. 
Defense performance is evaluated by certified accuracy (CA), following the setup of \citep{salman2020denoised, zhang2022robustify}. 
CA is defined using the $\ell_2$ norm-based input perturbation radius $r$, where a larger $r$ indicates a stronger adversarial threat.
We refer readers to Appx.\,\ref{app: black_box_defense} for more experiment details. 
\textbf{Tab.\,\ref{tab: zo_ds_r}} highlights that DeepZero consistently outperforms ZO-AE-DS in terms of CA for all values of $r > 0$. 
It is important to note that when $r = 0$, CA is equivalent to the standard testing accuracy. 
This indicates that DeepZero excels over ZO-AE-DS not only in adversarial robustness but also in overall generalization performance.

\begin{wrapfigure}{r}{48mm}
\vspace{-5mm}
\centerline{
\includegraphics[width=48mm,height=!]{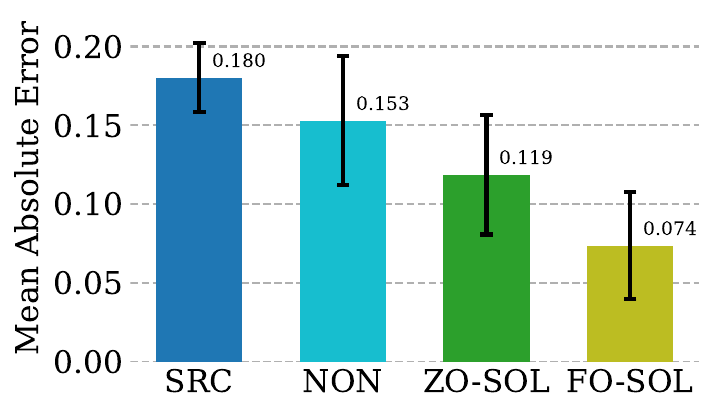}
}
\caption{\footnotesize{
Average MAE of corrected low-fidelity simulations compared to high-fidelity simulations over 5 test simulations using different correction methods. 
Error bar is variance of MAE across 5 test simulations.
}}
  \label{fig: sitl}
 \vspace*{-4.5mm}
\end{wrapfigure}
\textbf{Simulation-coupled DL for discretized PDE error correction.}
Numerical methods, while instrumental in providing physics-informed simulations, come with their own challenge: the discretization unavoidably produces numerical errors. 
DL has gained significant attention for addressing this error correction problem. 
The feasibility of training a corrective neural network through looping interactions with the iterative partial differential equation (PDE) solver, coined `solver-in-the-loop' (\textbf{SOL}), has been demonstrated in \citep{um2020solver}.
While existing work focused on using or developing differentiable simulators for model training, we extend SOL by leveraging {\DeepZero}, enabling its use with non-differentiable or black-box simulators. 
We name our method \textbf{ZO-SOL} and refer readers to \textbf{Fig.\,\ref{fig: sitl_methods_illustration}} in Appx.\,\ref{app: sitl}
for a schematic overview.
In this experimental framework, the goal is to correct the output of a low fidelity (\textit{i.e.}, coarse mesh) simulation using a learned DNN so that the corrected simulation more closely aligns with output of a  high fidelity (\textit{i.e.}, fine mesh) simulation.
In an effort to reduce the amount of correction required by DNN, this correction is applied at each simulation timestep and the corrected simulation state is provided to the simulator to compute the next timestep.
For our experiments, we consider 2D unsteady wake flow fluid dynamics benchmark from \citep{um2020solver}, in which there is a static rod around which fluid is flowing, and utilize PhiFlow \citep{holl2020learning} as the simulator.
Additional details on the PDE, the solver-in-the-loop loss, and the DNN architecture are given in Appx.\,\ref{app: sitl}.
\textbf{Fig.\,\ref{fig: sitl}} compares the test error correction performance of ZO-SOL (via DeepZero) with three differentiable approaches methods considered in \citep{um2020solver}: SRC (low fidelity simulation without error correction), NON (non-interactive training out of the simulation loop using pre-generated low and high fidelity simulation data), and FO-SOL (FO training for SOL given a differentiable simulator). 
The simulation error for each method in \textbf{Fig.\,\ref{fig: sitl}} is measured as the average across the five test simulations (with varying Reynold's numbers that were not utilized in the training set of simulation data). 
The error for each test simulation is computed as the mean absolute error (MAE) of the corrected simulation compared to the high fidelity simulation averaged across all simulation timesteps.
We implement DeepZero for ZO-SOL following the setup used in the image classification task, except for choosing   a 95\% gradient sparsity. 
The ZO-SOL and FO-SOL use 16 unrolling steps in the loss function to allow the correction function to interact with the simulator during training.
The results demonstrate that ZO-SOL achieved by DeepZero outperforms the SRC and NON baselines, and narrows the performance gap with FO-SOL, despite only having query-based access to the simulator. 
Comparing ZO-SOL with NON 
highlights the promise of ZO-SOL even when integrated with black-box simulators.

\vspace*{-5mm}
\section{Conclusion}
\vspace*{-6mm}

This paper introduces {\DeepZero}, a framework designed to enhance the scalability of ZO optimization for deep network training. Specifically, {\DeepZero} integrates coordinate-wise gradient estimation, ZO pruning-enabled gradient sparsity, feature reuse, and forward parallelization into a unified training pipeline. 
Leveraging these innovations, {\DeepZero} showcases both efficiency and effectiveness in a wide range of applications, including image classification tasks and various practical black-box DL scenarios. 
While {\DeepZero} has made remarkable progress in training DL models on datasets like ResNet-20 and CIFAR-10, it is important to acknowledge that scalability remains a significant challenge when dealing with even larger models and more extensive datasets. 
Future studies to accelerate ZO optimization for DL are necessary.  
Additionally, it is worthwhile to explore the applicability of {\DeepZero} in other domains, such as digital twin applications that  involve non-differentiable physical entities, and on-device training  where the computational overhead of establishing computational graphs and backpropagation is not feasible. 
Lastly, we refer readers to Appx.\,\ref{app: broader_impact} for a discussion of the broader impact of this paper.

\section*{Acknowledgment}
We thank the U.S. Department of Energy via Lawrence Livermore National Laboratory under Contract DE-AC52- 07NA27344 and the LLNL-LDRD Program under Project No. 23-ER-030 (LLNL-CONF-849161) for their support.

\bibliography{bibs/zo, bibs/pruning, bibs/misc, bibs/DFO, bibs/distributed}

\begin{thebibliography}{107}
\providecommand{\natexlab}[1]{#1}
\providecommand{\url}[1]{\texttt{#1}}
\expandafter\ifx\csname urlstyle\endcsname\relax
  \providecommand{\doi}[1]{doi: #1}\else
  \providecommand{\doi}{doi: \begingroup \urlstyle{rm}\Url}\fi

\bibitem[Abreu~de Souza et~al.(2023)Abreu~de Souza, Crispim Rom\~ao, Castro, Nikjoo, and Porod]{souza2023exploring}
Fernando Abreu~de Souza, Miguel Crispim Rom\~ao, Nuno~Filipe Castro, Mehraveh Nikjoo, and Werner Porod.
\newblock Exploring parameter spaces with artificial intelligence and machine learning black-box optimization algorithms.
\newblock \emph{Phys. Rev. D}, 107:\penalty0 035004, Feb 2023.
\newblock \doi{10.1103/PhysRevD.107.035004}.
\newblock URL \url{https://link.aps.org/doi/10.1103/PhysRevD.107.035004}.

\bibitem[Amari(1993)]{amari1993backpropagation}
Shun-ichi Amari.
\newblock Backpropagation and stochastic gradient descent method.
\newblock \emph{Neurocomputing}, 5\penalty0 (4-5):\penalty0 185--196, 1993.

\bibitem[Balasubramanian \& Ghadimi(2018)Balasubramanian and Ghadimi]{balasubramanian2018zeroth}
Krishnakumar Balasubramanian and Saeed Ghadimi.
\newblock Zeroth-order (non)-convex stochastic optimization via conditional gradient and gradient updates.
\newblock \emph{NeurIPS}, 31, 2018.

\bibitem[Bartoldson et~al.(2023)Bartoldson, Kailkhura, and Blalock]{bartoldson2023compute}
Brian~R Bartoldson, Bhavya Kailkhura, and Davis Blalock.
\newblock Compute-efficient deep learning: Algorithmic trends and opportunities.
\newblock \emph{Journal of Machine Learning Research}, 24:\penalty0 1--77, 2023.

\bibitem[Baydin et~al.(2020)Baydin, NYU, Feickert, Gray, Heinrich, NYU, Neubauer, Pearkes, Simpson, Smith, et~al.]{baydin2020differentiable}
At{\i}l{\i}m~G{\"u}nes Baydin, Kyle~Cranmer NYU, Matthew Feickert, Lindsey Gray, Lukas Heinrich, Alexander~Held NYU, Andrew Melo Vanderbilt~Mark Neubauer, Jannicke Pearkes, Nathan Simpson, Nick Smith, et~al.
\newblock Differentiable programming in high-energy physics.
\newblock \emph{Submitted as a Snowmass LOI}, 2020.

\bibitem[Baydin et~al.(2022)Baydin, Pearlmutter, Syme, Wood, and Torr]{baydin2022gradients}
At{\i}l{\i}m~G{\"u}ne{\c{s}} Baydin, Barak~A Pearlmutter, Don Syme, Frank Wood, and Philip Torr.
\newblock Gradients without backpropagation.
\newblock \emph{arXiv preprint arXiv:2202.08587}, 2022.

\bibitem[Belouze(2022)]{belouze2022optimization}
Gabriel Belouze.
\newblock Optimization without backpropagation.
\newblock \emph{arXiv preprint arXiv:2209.06302}, 2022.

\bibitem[Berahas et~al.(2022)Berahas, Cao, Choromanski, and Scheinberg]{berahas2022theoretical}
Albert~S Berahas, Liyuan Cao, Krzysztof Choromanski, and Katya Scheinberg.
\newblock A theoretical and empirical comparison of gradient approximations in derivative-free optimization.
\newblock \emph{Foundations of Computational Mathematics}, 22\penalty0 (2):\penalty0 507--560, 2022.

\bibitem[Blalock et~al.(2020)Blalock, Gonzalez~Ortiz, Frankle, and Guttag]{blalock2020state}
Davis Blalock, Jose~Javier Gonzalez~Ortiz, Jonathan Frankle, and John Guttag.
\newblock What is the state of neural network pruning?
\newblock \emph{Proceedings of machine learning and systems}, 2:\penalty0 129--146, 2020.

\bibitem[Boopathy \& Fiete(2022)Boopathy and Fiete]{boopathy2022train}
Akhilan Boopathy and Ila Fiete.
\newblock How to train your wide neural network without backprop: An input-weight alignment perspective.
\newblock In \emph{ICML}, pp.\  2178--2205. PMLR, 2022.

\bibitem[Bortz \& Kelley(1998)Bortz and Kelley]{bortz1998simplex}
David~Matthew Bortz and Carl~Tim Kelley.
\newblock The simplex gradient and noisy optimization problems.
\newblock In \emph{Computational Methods for Optimal Design and Control: Proceedings of the AFOSR Workshop on Optimal Design and Control Arlington, Virginia 30 September--3 October, 1997}, pp.\  77--90. Springer, 1998.

\bibitem[Bottou(2010)]{bottou2010large}
L{\'e}on Bottou.
\newblock Large-scale machine learning with stochastic gradient descent.
\newblock In \emph{Proceedings of COMPSTAT'2010: 19th International Conference on Computational StatisticsParis France, August 22-27, 2010 Keynote, Invited and Contributed Papers}, pp.\  177--186. Springer, 2010.

\bibitem[Bottou(2012)]{bottou2012stochastic}
L{\'e}on Bottou.
\newblock Stochastic gradient descent tricks.
\newblock \emph{Neural Networks: Tricks of the Trade: Second Edition}, pp.\  421--436, 2012.

\bibitem[Cai et~al.(2021)Cai, Lou, McKenzie, and Yin]{cai2021zeroth}
HanQin Cai, Yuchen Lou, Daniel McKenzie, and Wotao Yin.
\newblock A zeroth-order block coordinate descent algorithm for huge-scale black-box optimization.
\newblock \emph{arXiv preprint arXiv:2102.10707}, 2021.

\bibitem[Cai et~al.(2022)Cai, Mckenzie, Yin, and Zhang]{cai2022zeroth}
HanQin Cai, Daniel Mckenzie, Wotao Yin, and Zhenliang Zhang.
\newblock Zeroth-order regularized optimization (zoro): Approximately sparse gradients and adaptive sampling.
\newblock \emph{SIAM Journal on Optimization}, 32\penalty0 (2):\penalty0 687--714, 2022.

\bibitem[Chen et~al.(2017)Chen, Zhang, Sharma, Yi, and Hsieh]{chen2017zoo}
Pin-Yu Chen, Huan Zhang, Yash Sharma, Jinfeng Yi, and Cho-Jui Hsieh.
\newblock Zoo: Zeroth order optimization based black-box attacks to deep neural networks without training substitute models.
\newblock In \emph{Proceedings of the 10th ACM workshop on artificial intelligence and security}, pp.\  15--26, 2017.

\bibitem[Cheng et~al.(2021)Cheng, Wu, and Zhu]{cheng2021convergence}
Shuyu Cheng, Guoqiang Wu, and Jun Zhu.
\newblock On the convergence of prior-guided zeroth-order optimization algorithms.
\newblock \emph{Advances in Neural Information Processing Systems}, 34:\penalty0 14620--14631, 2021.

\bibitem[Chiang et~al.(2023)Chiang, Ni, Miller, Bansal, Geiping, Goldblum, and Goldstein]{chiang2023loss}
Ping-yeh Chiang, Renkun Ni, David~Yu Miller, Arpit Bansal, Jonas Geiping, Micah Goldblum, and Tom Goldstein.
\newblock Loss landscapes are all you need: Neural network generalization can be explained without the implicit bias of gradient descent.
\newblock In \emph{The Eleventh International Conference on Learning Representations}, 2023.

\bibitem[Cohen et~al.(2019)Cohen, Rosenfeld, and Kolter]{cohen2019certified}
Jeremy Cohen, Elan Rosenfeld, and Zico Kolter.
\newblock Certified adversarial robustness via randomized smoothing.
\newblock In \emph{international conference on machine learning}, pp.\  1310--1320. PMLR, 2019.

\bibitem[Conn et~al.(2009)Conn, Scheinberg, and Vicente]{conn2009introduction}
A.~R. Conn, K.~Scheinberg, and L.~N. Vicente.
\newblock \emph{Introduction to derivative-free optimization}, volume~8.
\newblock Siam, 2009.

\bibitem[Conn et~al.(2000)Conn, Gould, and Toint]{conn2000trust}
Andrew~R Conn, Nicholas~IM Gould, and Philippe~L Toint.
\newblock \emph{Trust region methods}.
\newblock SIAM, 2000.

\bibitem[Dhurandhar et~al.(2019)Dhurandhar, Pedapati, Balakrishnan, Chen, Shanmugam, and Puri]{dhurandhar2019model}
Amit Dhurandhar, Tejaswini Pedapati, Avinash Balakrishnan, Pin-Yu Chen, Karthikeyan Shanmugam, and Ruchir Puri.
\newblock Model agnostic contrastive explanations for structured data.
\newblock \emph{arXiv preprint arXiv:1906.00117}, 2019.

\bibitem[Diao et~al.(2022)Diao, Huang, Xu, Li, Lin, Zhou, and Zhang]{diao2022black}
Shizhe Diao, Zhichao Huang, Ruijia Xu, Xuechun Li, Yong Lin, Xiao Zhou, and Tong Zhang.
\newblock Black-box prompt learning for pre-trained language models.
\newblock \emph{arXiv preprint arXiv:2201.08531}, 2022.

\bibitem[Diffenderfer \& Kailkhura(2021)Diffenderfer and Kailkhura]{diffenderfer2021multi}
James Diffenderfer and Bhavya Kailkhura.
\newblock Multi-prize lottery ticket hypothesis: Finding accurate binary neural networks by pruning a randomly weighted network.
\newblock \emph{arXiv preprint arXiv:2103.09377}, 2021.

\bibitem[Diffenderfer et~al.(2021)Diffenderfer, Bartoldson, Chaganti, Zhang, and Kailkhura]{diffenderfer2021winning}
James Diffenderfer, Brian Bartoldson, Shreya Chaganti, Jize Zhang, and Bhavya Kailkhura.
\newblock A winning hand: Compressing deep networks can improve out-of-distribution robustness.
\newblock \emph{Advances in Neural Information Processing Systems}, 34:\penalty0 664--676, 2021.

\bibitem[Duchi et~al.(2015)Duchi, Jordan, Wainwright, and Wibisono]{duchi2015optimal}
John~C Duchi, Michael~I Jordan, Martin~J Wainwright, and Andre Wibisono.
\newblock Optimal rates for zero-order convex optimization: The power of two function evaluations.
\newblock \emph{IEEE Transactions on Information Theory}, 61\penalty0 (5):\penalty0 2788--2806, 2015.

\bibitem[Eriksson et~al.(2019)Eriksson, Pearce, Gardner, Turner, and Poloczek]{eriksson2019scalable}
David Eriksson, Michael Pearce, Jacob Gardner, Ryan~D Turner, and Matthias Poloczek.
\newblock Scalable global optimization via local bayesian optimization.
\newblock \emph{Advances in neural information processing systems}, 32, 2019.

\bibitem[Fang et~al.(2022)Fang, Liu, Zhang, Zhang, Ma, Li, Hu, Jiang, and Liu]{fang2022complex}
Yu~Fang, Jiancheng Liu, Mingrui Zhang, Jiasheng Zhang, Yidong Ma, Minchen Li, Yuanming Hu, Chenfanfu Jiang, and Tiantian Liu.
\newblock Complex locomotion skill learning via differentiable physics.
\newblock \emph{arXiv preprint arXiv:2206.02341}, 2022.

\bibitem[Fermi(1952)]{fermi1952numerical}
Enrico Fermi.
\newblock Numerical solution of a minimum problem.
\newblock Technical report, Los Alamos Scientific Lab., Los Alamos, NM, 1952.

\bibitem[Flaxman et~al.(2005)Flaxman, Kalai, and McMahan]{flaxman2005online}
A.~D. Flaxman, A.~T. Kalai, and H.~B. McMahan.
\newblock Online convex optimization in the bandit setting: {G}radient descent without a gradient.
\newblock In \emph{Proceedings of the sixteenth annual ACM-SIAM symposium on Discrete algorithms}, pp.\  385--394, 2005.

\bibitem[Frankle \& Carbin(2018)Frankle and Carbin]{frankle2018lottery}
Jonathan Frankle and Michael Carbin.
\newblock The lottery ticket hypothesis: Finding sparse, trainable neural networks.
\newblock \emph{arXiv preprint arXiv:1803.03635}, 2018.

\bibitem[Gardner(1984)]{gardner1984learning}
William~A Gardner.
\newblock Learning characteristics of stochastic-gradient-descent algorithms: A general study, analysis, and critique.
\newblock \emph{Signal processing}, 6\penalty0 (2):\penalty0 113--133, 1984.

\bibitem[Ghadimi \& Lan(2013)Ghadimi and Lan]{ghadimi2013stochastic}
S.~Ghadimi and G.~Lan.
\newblock Stochastic first-and zeroth-order methods for nonconvex stochastic programming.
\newblock \emph{SIAM Journal on Optimization}, 23\penalty0 (4):\penalty0 2341--2368, 2013.

\bibitem[Goyal et~al.(2017)Goyal, Doll{\'a}r, Girshick, Noordhuis, Wesolowski, Kyrola, Tulloch, Jia, and He]{goyal2017accurate}
Priya Goyal, Piotr Doll{\'a}r, Ross Girshick, Pieter Noordhuis, Lukasz Wesolowski, Aapo Kyrola, Andrew Tulloch, Yangqing Jia, and Kaiming He.
\newblock Accurate, large minibatch sgd: Training imagenet in 1 hour.
\newblock \emph{arXiv preprint arXiv:1706.02677}, 2017.

\bibitem[Greengard(2020)]{greengard2020neuromorphic}
Samuel Greengard.
\newblock Neuromorphic chips take shape.
\newblock \emph{Communications of the ACM}, 63\penalty0 (8):\penalty0 9--11, 2020.

\bibitem[Grefenstette(1993)]{grefenstette1993genetic}
John~J Grefenstette.
\newblock Genetic algorithms and machine learning.
\newblock In \emph{Proceedings of the sixth annual conference on Computational learning theory}, pp.\  3--4, 1993.

\bibitem[Gu et~al.(2021{\natexlab{a}})Gu, Liu, Zhang, Geng, and Huang]{gu2021optimizing}
Bin Gu, Guodong Liu, Yanfu Zhang, Xiang Geng, and Heng Huang.
\newblock Optimizing large-scale hyperparameters via automated learning algorithm.
\newblock \emph{arXiv preprint arXiv:2102.09026}, 2021{\natexlab{a}}.

\bibitem[Gu et~al.(2021{\natexlab{b}})Gu, Feng, Zhao, Ying, Chen, and Pan]{gu2021efficient}
Jiaqi Gu, Chenghao Feng, Zheng Zhao, Zhoufeng Ying, Ray~T Chen, and David~Z Pan.
\newblock Efficient on-chip learning for optical neural networks through power-aware sparse zeroth-order optimization.
\newblock In \emph{Proceedings of the AAAI Conference on Artificial Intelligence}, volume~35, pp.\  7583--7591, 2021{\natexlab{b}}.

\bibitem[Gu et~al.(2021{\natexlab{c}})Gu, Zhu, Feng, Jiang, Chen, and Pan]{gu2021l2ight}
Jiaqi Gu, Hanqing Zhu, Chenghao Feng, Zixuan Jiang, Ray Chen, and David Pan.
\newblock L2ight: Enabling on-chip learning for optical neural networks via efficient in-situ subspace optimization.
\newblock \emph{Advances in Neural Information Processing Systems}, 34:\penalty0 8649--8661, 2021{\natexlab{c}}.

\bibitem[Han et~al.(2015)Han, Mao, and Dally]{han2015deep}
Song Han, Huizi Mao, and William~J Dally.
\newblock Deep compression: Compressing deep neural networks with pruning, trained quantization and huffman coding.
\newblock \emph{arXiv preprint arXiv:1510.00149}, 2015.

\bibitem[Hinton(2022)]{hinton2022forward}
Geoffrey Hinton.
\newblock The forward-forward algorithm: Some preliminary investigations.
\newblock \emph{arXiv preprint arXiv:2212.13345}, 2022.

\bibitem[Hogan \& Kailkhura(2018)Hogan and Kailkhura]{hogan2018universal}
Thomas~A Hogan and Bhavya Kailkhura.
\newblock Universal decision-based black-box perturbations: Breaking security-through-obscurity defenses.
\newblock \emph{arXiv preprint arXiv:1811.03733}, 2018.

\bibitem[Holl et~al.(2020)Holl, Koltun, and Thuerey]{holl2020learning}
Philipp Holl, Vladlen Koltun, and Nils Thuerey.
\newblock Learning to control pdes with differentiable physics, 2020.

\bibitem[Hu et~al.(2019)Hu, Liu, Spielberg, Tenenbaum, Freeman, Wu, Rus, and Matusik]{hu2019chainqueen}
Yuanming Hu, Jiancheng Liu, Andrew Spielberg, Joshua~B Tenenbaum, William~T Freeman, Jiajun Wu, Daniela Rus, and Wojciech Matusik.
\newblock Chainqueen: A real-time differentiable physical simulator for soft robotics.
\newblock In \emph{ICRA}, pp.\  6265--6271. IEEE, 2019.

\bibitem[Ilyas et~al.(2018{\natexlab{a}})Ilyas, Engstrom, Athalye, and Lin]{ilyas2018black}
Andrew Ilyas, Logan Engstrom, Anish Athalye, and Jessy Lin.
\newblock Black-box adversarial attacks with limited queries and information.
\newblock In \emph{International conference on machine learning}, pp.\  2137--2146. PMLR, 2018{\natexlab{a}}.

\bibitem[Ilyas et~al.(2018{\natexlab{b}})Ilyas, Engstrom, and Madry]{ilyas2018prior}
Andrew Ilyas, Logan Engstrom, and Aleksander Madry.
\newblock Prior convictions: Black-box adversarial attacks with bandits and priors.
\newblock \emph{arXiv preprint arXiv:1807.07978}, 2018{\natexlab{b}}.

\bibitem[Isomura \& Toyoizumi(2018)Isomura and Toyoizumi]{isomura2018error}
Takuya Isomura and Taro Toyoizumi.
\newblock Error-gated hebbian rule: A local learning rule for principal and independent component analysis.
\newblock \emph{Scientific reports}, 8\penalty0 (1):\penalty0 1835, 2018.

\bibitem[Jabri \& Flower(1992)Jabri and Flower]{jabri1992weight}
Marwan Jabri and Barry Flower.
\newblock Weight perturbation: An optimal architecture and learning technique for analog vlsi feedforward and recurrent multilayer networks.
\newblock \emph{IEEE Transactions on Neural Networks}, 3\penalty0 (1):\penalty0 154--157, 1992.

\bibitem[Jaderberg et~al.(2017)Jaderberg, Czarnecki, Osindero, Vinyals, Graves, Silver, and Kavukcuoglu]{jaderberg2017decoupled}
Max Jaderberg, Wojciech~Marian Czarnecki, Simon Osindero, Oriol Vinyals, Alex Graves, David Silver, and Koray Kavukcuoglu.
\newblock Decoupled neural interfaces using synthetic gradients.
\newblock In \emph{International conference on machine learning}, pp.\  1627--1635. PMLR, 2017.

\bibitem[Kiefer \& Wolfowitz(1952)Kiefer and Wolfowitz]{kiefer1952stochastic}
Jack Kiefer and Jacob Wolfowitz.
\newblock Stochastic estimation of the maximum of a regression function.
\newblock \emph{The Annals of Mathematical Statistics}, pp.\  462--466, 1952.

\bibitem[Kingma \& Ba(2014)Kingma and Ba]{kingma2014adam}
Diederik~P Kingma and Jimmy Ba.
\newblock Adam: A method for stochastic optimization.
\newblock \emph{arXiv preprint arXiv:1412.6980}, 2014.

\bibitem[Larson et~al.(2019)Larson, Menickelly, and Wild]{larson2019derivative}
Jeffrey Larson, Matt Menickelly, and Stefan~M Wild.
\newblock Derivative-free optimization methods.
\newblock \emph{Acta Numerica}, 28:\penalty0 287--404, 2019.

\bibitem[Lee et~al.(2018)Lee, Ajanthan, and Torr]{lee2018snip}
Namhoon Lee, Thalaiyasingam Ajanthan, and Philip~HS Torr.
\newblock Snip: Single-shot network pruning based on connection sensitivity.
\newblock \emph{arXiv preprint arXiv:1810.02340}, 2018.

\bibitem[Lian et~al.(2016)Lian, Zhang, Hsieh, Huang, and Liu]{lian2016comprehensive}
X.~Lian, H.~Zhang, C.-J. Hsieh, Y.~Huang, and J.~Liu.
\newblock A comprehensive linear speedup analysis for asynchronous stochastic parallel optimization from zeroth-order to first-order.
\newblock In \emph{Advances in Neural Information Processing Systems}, pp.\  3054--3062, 2016.

\bibitem[Lillicrap et~al.(2014)Lillicrap, Cownden, Tweed, and Akerman]{lillicrap2014random}
Timothy~P Lillicrap, Daniel Cownden, Douglas~B Tweed, and Colin~J Akerman.
\newblock Random feedback weights support learning in deep neural networks.
\newblock \emph{arXiv preprint arXiv:1411.0247}, 2014.

\bibitem[Liu et~al.(2018{\natexlab{a}})Liu, Chen, Chen, and Hero]{liu2018zeroth_admm}
Sijia Liu, Jie Chen, Pin-Yu Chen, and Alfred Hero.
\newblock Zeroth-order online alternating direction method of multipliers: Convergence analysis and applications.
\newblock In \emph{International Conference on Artificial Intelligence and Statistics}, pp.\  288--297. PMLR, 2018{\natexlab{a}}.

\bibitem[Liu et~al.(2018{\natexlab{b}})Liu, Kailkhura, Chen, Ting, Chang, and Amini]{liu2018zeroth}
Sijia Liu, Bhavya Kailkhura, Pin-Yu Chen, Paishun Ting, Shiyu Chang, and Lisa Amini.
\newblock Zeroth-order stochastic variance reduction for nonconvex optimization.
\newblock \emph{Advances in Neural Information Processing Systems}, 31, 2018{\natexlab{b}}.

\bibitem[Liu et~al.(2020{\natexlab{a}})Liu, Chen, Kailkhura, Zhang, Hero~III, and Varshney]{liu2020primer}
Sijia Liu, Pin-Yu Chen, Bhavya Kailkhura, Gaoyuan Zhang, Alfred~O Hero~III, and Pramod~K Varshney.
\newblock A primer on zeroth-order optimization in signal processing and machine learning: Principals, recent advances, and applications.
\newblock \emph{IEEE Signal Processing Magazine}, 37\penalty0 (5):\penalty0 43--54, 2020{\natexlab{a}}.

\bibitem[Liu et~al.(2020{\natexlab{b}})Liu, Lu, Chen, Feng, Xu, Al-Dujaili, Hong, and O’Reilly]{liu2020min}
Sijia Liu, Songtao Lu, Xiangyi Chen, Yao Feng, Kaidi Xu, Abdullah Al-Dujaili, Mingyi Hong, and Una-May O’Reilly.
\newblock Min-max optimization without gradients: Convergence and applications to black-box evasion and poisoning attacks.
\newblock In \emph{International conference on machine learning}, pp.\  6282--6293. PMLR, 2020{\natexlab{b}}.

\bibitem[Louppe et~al.(2019)Louppe, Hermans, and Cranmer]{louppe2019adversarial}
Gilles Louppe, Joeri Hermans, and Kyle Cranmer.
\newblock Adversarial variational optimization of non-differentiable simulators.
\newblock In \emph{The 22nd International Conference on Artificial Intelligence and Statistics}, pp.\  1438--1447. PMLR, 2019.

\bibitem[Ma et~al.(2021)Ma, Yuan, Shen, Chen, Chen, Chen, Liu, Qin, Liu, Wang, et~al.]{ma2021sanity}
Xiaolong Ma, Geng Yuan, Xuan Shen, Tianlong Chen, Xuxi Chen, Xiaohan Chen, Ning Liu, Minghai Qin, Sijia Liu, Zhangyang Wang, et~al.
\newblock Sanity checks for lottery tickets: Does your winning ticket really win the jackpot?
\newblock \emph{Advances in Neural Information Processing Systems}, 34:\penalty0 12749--12760, 2021.

\bibitem[Madry et~al.(2017)Madry, Makelov, Schmidt, Tsipras, and Vladu]{madry2017towards}
Aleksander Madry, Aleksandar Makelov, Ludwig Schmidt, Dimitris Tsipras, and Adrian Vladu.
\newblock Towards deep learning models resistant to adversarial attacks.
\newblock \emph{arXiv preprint arXiv:1706.06083}, 2017.

\bibitem[Malladi et~al.(2023)Malladi, Gao, Nichani, Damian, Lee, Chen, and Arora]{malladi2023fine}
Sadhika Malladi, Tianyu Gao, Eshaan Nichani, Alex Damian, Jason~D Lee, Danqi Chen, and Sanjeev Arora.
\newblock Fine-tuning language models with just forward passes.
\newblock \emph{arXiv preprint arXiv:2305.17333}, 2023.

\bibitem[Meier et~al.(2019)Meier, Mujika, Gauy, and Steger]{meier2019improving}
Florian Meier, Asier Mujika, Marcelo~Matheus Gauy, and Angelika Steger.
\newblock Improving gradient estimation in evolutionary strategies with past descent directions.
\newblock \emph{arXiv preprint arXiv:1910.05268}, 2019.

\bibitem[Moraitis et~al.(2022)Moraitis, Toichkin, Journ{\'e}, Chua, and Guo]{moraitis2022softhebb}
Timoleon Moraitis, Dmitry Toichkin, Adrien Journ{\'e}, Yansong Chua, and Qinghai Guo.
\newblock Softhebb: Bayesian inference in unsupervised hebbian soft winner-take-all networks.
\newblock \emph{Neuromorphic Computing and Engineering}, 2\penalty0 (4):\penalty0 044017, 2022.

\bibitem[Nelder \& Mead(1965)Nelder and Mead]{nelder1965simplex}
John~A Nelder and Roger Mead.
\newblock A simplex method for function minimization.
\newblock \emph{The computer journal}, 7\penalty0 (4):\penalty0 308--313, 1965.

\bibitem[Nesterov \& Spokoiny(2015)Nesterov and Spokoiny]{nesterov2015random}
Y.~Nesterov and V.~Spokoiny.
\newblock Random gradient-free minimization of convex functions.
\newblock \emph{Foundations of Computational Mathematics}, 2\penalty0 (17):\penalty0 527--566, 2015.

\bibitem[Nesterov \& Spokoiny(2017)Nesterov and Spokoiny]{nesterov2017random}
Yurii Nesterov and Vladimir Spokoiny.
\newblock Random gradient-free minimization of convex functions.
\newblock \emph{Foundations of Computational Mathematics}, 17:\penalty0 527--566, 2017.

\bibitem[N{\o}kland(2016)]{nokland2016direct}
Arild N{\o}kland.
\newblock Direct feedback alignment provides learning in deep neural networks.
\newblock \emph{Advances in neural information processing systems}, 29, 2016.

\bibitem[N{\o}kland \& Eidnes(2019)N{\o}kland and Eidnes]{nokland2019training}
Arild N{\o}kland and Lars~Hiller Eidnes.
\newblock Training neural networks with local error signals.
\newblock In \emph{International conference on machine learning}, pp.\  4839--4850. PMLR, 2019.

\bibitem[Ohta et~al.(2020)Ohta, Berger, Sokolov, and Riezler]{ohta2020sparse}
Mayumi Ohta, Nathaniel Berger, Artem Sokolov, and Stefan Riezler.
\newblock Sparse perturbations for improved convergence in stochastic zeroth-order optimization.
\newblock In \emph{Machine Learning, Optimization, and Data Science: 6th International Conference, LOD 2020, Siena, Italy, July 19--23, 2020, Revised Selected Papers, Part II 6}, pp.\  39--64. Springer, 2020.

\bibitem[Ren et~al.(2022)Ren, Kornblith, Liao, and Hinton]{ren2022scaling}
Mengye Ren, Simon Kornblith, Renjie Liao, and Geoffrey Hinton.
\newblock Scaling forward gradient with local losses.
\newblock \emph{arXiv preprint arXiv:2210.03310}, 2022.

\bibitem[Rios \& Sahinidis(2013)Rios and Sahinidis]{rios2013derivative}
L.~M. Rios and N.~V. Sahinidis.
\newblock Derivative-free optimization: a review of algorithms and comparison of software implementations.
\newblock \emph{Journal of Global Optimization}, 56\penalty0 (3):\penalty0 1247--1293, 2013.

\bibitem[Rumelhart et~al.(1995)Rumelhart, Durbin, Golden, and Chauvin]{rumelhart1995backpropagation}
David~E Rumelhart, Richard Durbin, Richard Golden, and Yves Chauvin.
\newblock Backpropagation: The basic theory.
\newblock \emph{Backpropagation: Theory, architectures and applications}, pp.\  1--34, 1995.

\bibitem[Salman et~al.(2020)Salman, Sun, Yang, Kapoor, and Kolter]{salman2020denoised}
Hadi Salman, Mingjie Sun, Greg Yang, Ashish Kapoor, and J~Zico Kolter.
\newblock Denoised smoothing: A provable defense for pretrained classifiers.
\newblock \emph{Advances in Neural Information Processing Systems}, 33:\penalty0 21945--21957, 2020.

\bibitem[Shahriari et~al.(2015)Shahriari, Swersky, Wang, Adams, and De~Freitas]{shahriari2015taking}
Bobak Shahriari, Kevin Swersky, Ziyu Wang, Ryan~P Adams, and Nando De~Freitas.
\newblock Taking the human out of the loop: A review of bayesian optimization.
\newblock \emph{Proceedings of the IEEE}, 104\penalty0 (1):\penalty0 148--175, 2015.

\bibitem[Shamir(2013)]{shamir2013complexity}
O.~Shamir.
\newblock On the complexity of bandit and derivative-free stochastic convex optimization.
\newblock In \emph{Conference on Learning Theory}, pp.\  3--24, 2013.

\bibitem[Shu et~al.(2022)Shu, Dai, Sng, Verma, Jaillet, and Low]{shu2022zeroth}
Yao Shu, Zhongxiang Dai, Weicong Sng, Arun Verma, Patrick Jaillet, and Bryan Kian~Hsiang Low.
\newblock Zeroth-order optimization with trajectory-informed derivative estimation.
\newblock In \emph{ICLR}, 2022.

\bibitem[Silver et~al.(2021)Silver, Goyal, Danihelka, Hessel, and van Hasselt]{silver2021learning}
David Silver, Anirudh Goyal, Ivo Danihelka, Matteo Hessel, and Hado van Hasselt.
\newblock Learning by directional gradient descent.
\newblock In \emph{International Conference on Learning Representations}, 2021.

\bibitem[Spall(1992)]{spall1992multivariate}
James~C Spall.
\newblock Multivariate stochastic approximation using a simultaneous perturbation gradient approximation.
\newblock \emph{IEEE transactions on automatic control}, 37\penalty0 (3):\penalty0 332--341, 1992.

\bibitem[Su et~al.(2020)Su, Chen, Cai, Wu, Gao, Wang, and Lee]{su2020sanity}
Jingtong Su, Yihang Chen, Tianle Cai, Tianhao Wu, Ruiqi Gao, Liwei Wang, and Jason~D Lee.
\newblock Sanity-checking pruning methods: Random tickets can win the jackpot.
\newblock \emph{Advances in Neural Information Processing Systems}, 33:\penalty0 20390--20401, 2020.

\bibitem[Sun et~al.(2022)Sun, Shao, Qian, Huang, and Qiu]{sun2022black}
Tianxiang Sun, Yunfan Shao, Hong Qian, Xuanjing Huang, and Xipeng Qiu.
\newblock Black-box tuning for language-model-as-a-service.
\newblock In \emph{International Conference on Machine Learning}, pp.\  20841--20855. PMLR, 2022.

\bibitem[Tanaka et~al.(2020)Tanaka, Kunin, Yamins, and Ganguli]{tanaka2020pruning}
Hidenori Tanaka, Daniel Kunin, Daniel~L Yamins, and Surya Ganguli.
\newblock Pruning neural networks without any data by iteratively conserving synaptic flow.
\newblock \emph{Advances in neural information processing systems}, 33:\penalty0 6377--6389, 2020.

\bibitem[Tavanaei et~al.(2019)Tavanaei, Ghodrati, Kheradpisheh, Masquelier, and Maida]{tavanaei2019deep}
Amirhossein Tavanaei, Masoud Ghodrati, Saeed~Reza Kheradpisheh, Timoth{\'e}e Masquelier, and Anthony Maida.
\newblock Deep learning in spiking neural networks.
\newblock \emph{Neural networks}, 111:\penalty0 47--63, 2019.

\bibitem[Thelen et~al.(2022)Thelen, Zhang, Fink, Lu, Ghosh, Youn, Todd, Mahadevan, Hu, and Hu]{thelen2022comprehensive}
Adam Thelen, Xiaoge Zhang, Olga Fink, Yan Lu, Sayan Ghosh, Byeng~D Youn, Michael~D Todd, Sankaran Mahadevan, Chao Hu, and Zhen Hu.
\newblock A comprehensive review of digital twin—part 1: modeling and twinning enabling technologies.
\newblock \emph{Structural and Multidisciplinary Optimization}, 65\penalty0 (12):\penalty0 354, 2022.

\bibitem[Torczon(1991)]{torczon1991convergence}
Virginia Torczon.
\newblock On the convergence of the multidirectional search algorithm.
\newblock \emph{SIAM journal on Optimization}, 1\penalty0 (1):\penalty0 123--145, 1991.

\bibitem[Tsai et~al.(2020)Tsai, Chen, and Ho]{tsai2020transfer}
Yun-Yun Tsai, Pin-Yu Chen, and Tsung-Yi Ho.
\newblock Transfer learning without knowing: Reprogramming black-box machine learning models with scarce data and limited resources.
\newblock In \emph{International Conference on Machine Learning}, pp.\  9614--9624. PMLR, 2020.

\bibitem[Tsaknakis et~al.(2022)Tsaknakis, Kailkhura, Liu, Loveland, Diffenderfer, Hiszpanski, and Hong]{tsaknakis2022zeroth}
Ioannis Tsaknakis, Bhavya Kailkhura, Sijia Liu, Donald Loveland, James Diffenderfer, Anna~Maria Hiszpanski, and Mingyi Hong.
\newblock Zeroth-order sciml: Non-intrusive integration of scientific software with deep learning.
\newblock \emph{arXiv preprint arXiv:2206.02785}, 2022.

\bibitem[Tu et~al.(2019)Tu, Ting, Chen, Liu, Zhang, Yi, Hsieh, and Cheng]{tu2019autozoom}
Chun-Chen Tu, Paishun Ting, Pin-Yu Chen, Sijia Liu, Huan Zhang, Jinfeng Yi, Cho-Jui Hsieh, and Shin-Ming Cheng.
\newblock Autozoom: Autoencoder-based zeroth order optimization method for attacking black-box neural networks.
\newblock In \emph{Proceedings of the AAAI Conference on Artificial Intelligence}, pp.\  742--749, 2019.

\bibitem[Um et~al.(2020)Um, Brand, Fei, Holl, and Thuerey]{um2020solver}
Kiwon Um, Robert Brand, Yun~Raymond Fei, Philipp Holl, and Nils Thuerey.
\newblock Solver-in-the-loop: Learning from differentiable physics to interact with iterative pde-solvers.
\newblock \emph{NeurIPS}, 33:\penalty0 6111--6122, 2020.

\bibitem[Urruty \& Lemar{\'e}chal(1993)Urruty and Lemar{\'e}chal]{urruty1993convex}
Jean-Baptiste~Hiriart Urruty and Claude Lemar{\'e}chal.
\newblock \emph{Convex analysis and minimization algorithms}.
\newblock Springer-Verlag, 1993.

\bibitem[Vaz \& Vicente(2009)Vaz and Vicente]{vaz2009pswarm}
A~Ismael~F Vaz and Lu{\'\i}s~Nunes Vicente.
\newblock Pswarm: a hybrid solver for linearly constrained global derivative-free optimization.
\newblock \emph{Optimization Methods \& Software}, 24\penalty0 (4-5):\penalty0 669--685, 2009.

\bibitem[Vemula et~al.(2019)Vemula, Sun, and Bagnell]{vemula2019contrasting}
Anirudh Vemula, Wen Sun, and J~Bagnell.
\newblock Contrasting exploration in parameter and action space: A zeroth-order optimization perspective.
\newblock In \emph{The 22nd International Conference on Artificial Intelligence and Statistics}, pp.\  2926--2935. PMLR, 2019.

\bibitem[Verma et~al.(2023)Verma, Bangar, Subramanyam, Lal, Shah, and Satoh]{verma2023certified}
Astha Verma, Siddhesh Bangar, AV~Subramanyam, Naman Lal, Rajiv~Ratn Shah, and Shin'ichi Satoh.
\newblock Certified zeroth-order black-box defense with robust unet denoiser.
\newblock \emph{arXiv preprint arXiv:2304.06430}, 2023.

\bibitem[Vicol et~al.(2023)Vicol, Kolter, and Swersky]{vicol2023low}
Paul Vicol, Zico Kolter, and Kevin Swersky.
\newblock Low-variance gradient estimation in unrolled computation graphs with es-single.
\newblock \emph{arXiv preprint arXiv:2304.11153}, 2023.

\bibitem[Wang et~al.(2020)Wang, Zhang, and Grosse]{wang2020picking}
Chaoqi Wang, Guodong Zhang, and Roger Grosse.
\newblock Picking winning tickets before training by preserving gradient flow.
\newblock \emph{arXiv preprint arXiv:2002.07376}, 2020.

\bibitem[Wang et~al.(2022)Wang, Guo, Su, Yang, and Yan]{wang2022zarts}
Xiaoxing Wang, Wenxuan Guo, Jianlin Su, Xiaokang Yang, and Junchi Yan.
\newblock Zarts: On zero-order optimization for neural architecture search.
\newblock \emph{Advances in Neural Information Processing Systems}, 35:\penalty0 12868--12880, 2022.

\bibitem[Wang et~al.(2017)Wang, Du, Balakrishnan, and Singh]{wang2017stochastic}
Yining Wang, Simon Du, Sivaraman Balakrishnan, and Aarti Singh.
\newblock Stochastic zeroth-order optimization in high dimensions.
\newblock \emph{arXiv preprint arXiv:1710.10551}, 2017.

\bibitem[Wright et~al.(1999)Wright, Nocedal, et~al.]{wright1999numerical}
Stephen Wright, Jorge Nocedal, et~al.
\newblock Numerical optimization.
\newblock \emph{Springer Science}, 35\penalty0 (67-68):\penalty0 7, 1999.

\bibitem[Ye et~al.(2018)Ye, Huang, Fang, Li, and Zhang]{ye2018hessian}
Haishan Ye, Zhichao Huang, Cong Fang, Chris~Junchi Li, and Tong Zhang.
\newblock Hessian-aware zeroth-order optimization for black-box adversarial attack.
\newblock \emph{arXiv preprint arXiv:1812.11377}, 2018.

\bibitem[You et~al.(2018)You, Zhang, Hsieh, Demmel, and Keutzer]{you2018imagenet}
Yang You, Zhao Zhang, Cho-Jui Hsieh, James Demmel, and Kurt Keutzer.
\newblock Imagenet training in minutes.
\newblock In \emph{Proceedings of the 47th International Conference on Parallel Processing}, pp.\  1--10, 2018.

\bibitem[Zhang et~al.(2017)Zhang, Zuo, Chen, Meng, and Zhang]{zhang2017beyond}
Kai Zhang, Wangmeng Zuo, Yunjin Chen, Deyu Meng, and Lei Zhang.
\newblock Beyond a gaussian denoiser: Residual learning of deep cnn for image denoising.
\newblock \emph{IEEE transactions on image processing}, 26\penalty0 (7):\penalty0 3142--3155, 2017.

\bibitem[Zhang et~al.(2022{\natexlab{a}})Zhang, Yao, Ram, Zhao, Chen, Hong, Wang, and Liu]{zhang2022advancing}
Yihua Zhang, Yuguang Yao, Parikshit Ram, Pu~Zhao, Tianlong Chen, Mingyi Hong, Yanzhi Wang, and Sijia Liu.
\newblock Advancing model pruning via bi-level optimization.
\newblock In \emph{Advances in Neural Information Processing Systems}, 2022{\natexlab{a}}.

\bibitem[Zhang et~al.(2024)Zhang, Li, Hong, Li, Zhang, Zheng, Chen, Lee, Yin, Hong, et~al.]{zhang2024revisiting}
Yihua Zhang, Pingzhi Li, Junyuan Hong, Jiaxiang Li, Yimeng Zhang, Wenqing Zheng, Pin-Yu Chen, Jason~D Lee, Wotao Yin, Mingyi Hong, et~al.
\newblock Revisiting zeroth-order optimization for memory-efficient llm fine-tuning: A benchmark.
\newblock \emph{arXiv preprint arXiv:2402.11592}, 2024.

\bibitem[Zhang et~al.(2022{\natexlab{b}})Zhang, Kamath, Wu, Fan, Chen, Wang, Chang, Liu, and Hao]{zhang2022data}
Yimeng Zhang, Akshay~Karkal Kamath, Qiucheng Wu, Zhiwen Fan, Wuyang Chen, Zhangyang Wang, Shiyu Chang, Sijia Liu, and Cong Hao.
\newblock Data-model-hardware tri-design for energy-efficient video intelligence.
\newblock \emph{arXiv preprint arXiv:2210.08578}, 2022{\natexlab{b}}.

\bibitem[Zhang et~al.(2022{\natexlab{c}})Zhang, Yao, Jia, Yi, Hong, Chang, and Liu]{zhang2022robustify}
Yimeng Zhang, Yuguang Yao, Jinghan Jia, Jinfeng Yi, Mingyi Hong, Shiyu Chang, and Sijia Liu.
\newblock How to robustify black-box ml models? a zeroth-order optimization perspective.
\newblock \emph{ICLR}, 2022{\natexlab{c}}.

\bibitem[Zhao et~al.(2019)Zhao, Liu, Chen, Hoang, Xu, Kailkhura, and Lin]{zhao2019design}
Pu~Zhao, Sijia Liu, Pin-Yu Chen, Nghia Hoang, Kaidi Xu, Bhavya Kailkhura, and Xue Lin.
\newblock On the design of black-box adversarial examples by leveraging gradient-free optimization and operator splitting method.
\newblock In \emph{Proceedings of the IEEE/CVF International Conference on Computer Vision}, pp.\  121--130, 2019.

\end{thebibliography}
\bibliographystyle{iclr2024_conference}

\appendix
\onecolumn
\setcounter{section}{0}

\section*{Appendix}

\setcounter{section}{0}
\setcounter{figure}{0}
\makeatletter 
\renewcommand{\thefigure}{A\arabic{figure}}
\renewcommand{\theHfigure}{A\arabic{figure}}
\renewcommand{\thetable}{A\arabic{table}}
\renewcommand{\theHtable}{A\arabic{table}}

\makeatother
\setcounter{table}{0}

\setcounter{mylemma}{0}
\renewcommand{\themylemma}{A\arabic{mylemma}}
\setcounter{equation}{0}
\renewcommand{\theequation}{A\arabic{equation}}

\section{Remark on convergence rate.}
\label{app: convergence_rate}
For a rigorous convergence analysis, we can perceive the sparsity-enhanced ZO training method introduced above as a specific instantiation within the broader framework of ZO stochastic coordinate descent \citep{lian2016comprehensive}. 
Thus, under necessary theoretical assumptions regarding Lipschitz continuity, gradient smoothness, and bounded gradient variance, we can obtain an upper bound for the convergence rate of our proposal in terms of the the gradient norm, serving as a stationarity-based convergence measure for non-convex optimization:

\vspace*{-5mm}
{\small
\begin{align}
\frac{\sum_{k=0}^K\mathbb{E}|\nabla f(\boldsymbol \theta_k)|^2}{K}\le O\left(\frac{1}{K(1-p)}+\sqrt{\frac{1}{K(1-p)}}\sigma\right),   
\end{align}
}%
where $k$ represents the iteration index, $K$ signifies the total number of iterations, $O$ is used in the context of big O notation, $p$ denotes the gradient sparsity ratio, and $\sigma$ stands for the upper bound on the variance of the stochastic gradients. The above highlights that as $p$ increases, the convergence error of the proposed approach also grows. However, it's important to note that increased sparsity leads to a reduced number of function queries. Consequently, we can see a provable tradeoff between the convergence error and the query efficiency, aligning well with our  understanding.

\section{The Simple CNN Architecture Considered for  Training w/ {\CGE} vs. {\RGE} }
\label{app: cge_vs_rge_cnn}

\begin{figure}[htb]
\centerline{\includegraphics[width=0.9\textwidth]{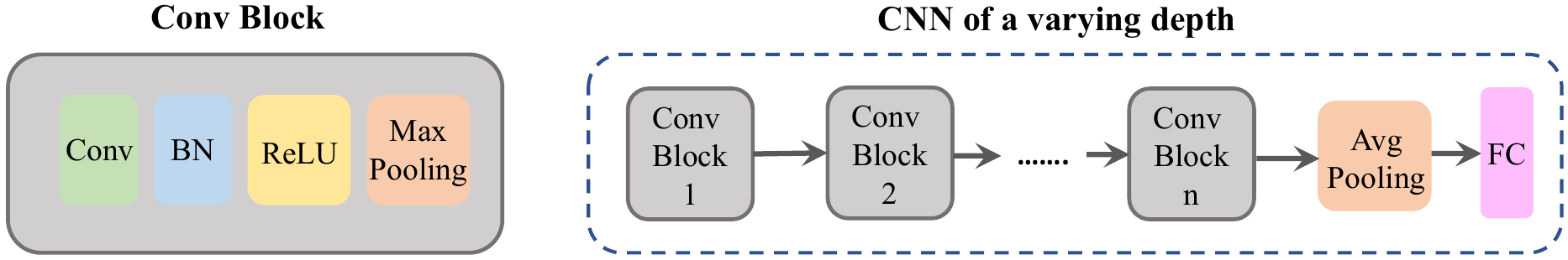}}
\caption{ \footnotesize{Illustration of the simple CNN considered with different depths. 
}}
\label{fig:cnn_arch}
\end{figure}

As illustrated in \textbf{Fig.\,\ref{fig:cnn_arch}}, the depth-adjustable simple CNN architecture comprises multiple conv blocks, an average pooling layer, and a fully-connected layer. Each conv block consists of four distinct layers,  including \ding{172} convolutional layer, \ding{173} batch normalization layer, \ding{174} ReLU layer, \ding{175} max pooling layer. 
In our experiments, the simple CNN networks are trained over a course of 50 epochs utilizing three distinct optimization strategies:
FO (first-order) SGD (stochastic gradient descent) training, ZO {\RGE}-based SGD training, and ZO {\CGE}-based  SGD training, together with a cosine learning rate scheduler is used.

\section{Computation Time Comparison between {\RGE} and {\CGE}}
\label{app: zo}
\begin{wrapfigure}{r}{50mm}
\vspace*{-4mm}
\centerline{
\includegraphics[width=50mm,height=!]{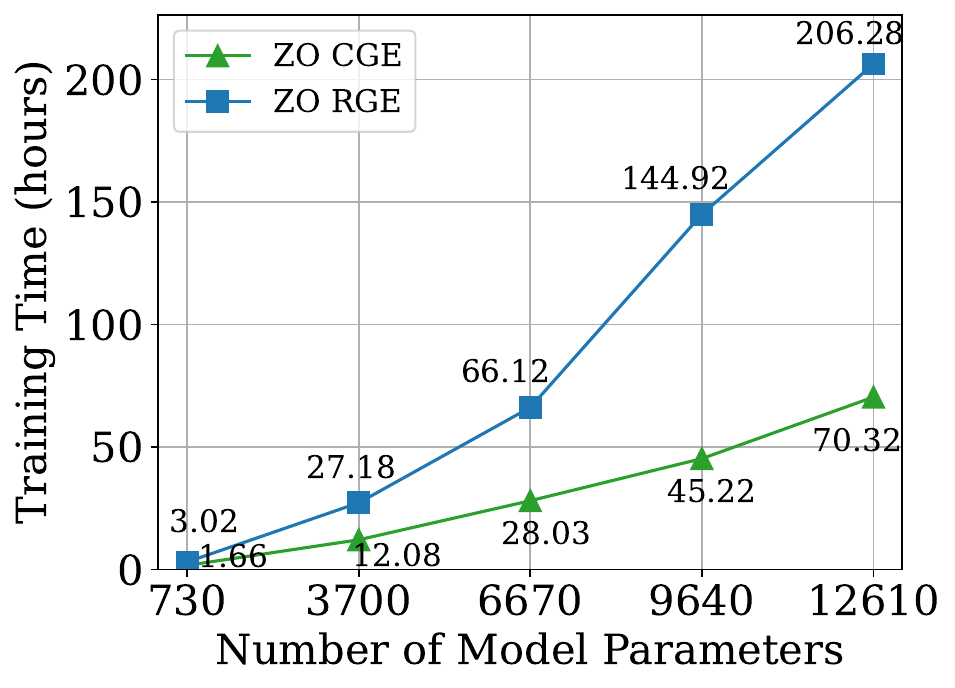}
}
\vspace*{-2mm}
\caption{\footnotesize{
The computation cost of using {\CGE} and {\RGE} to train the simple CNN in Fig.\,\ref{fig: Acc_CGE_RGE} over 50 epochs on CIFAR-10  against parameter size. 
}}
\vspace*{-5mm}
\label{fig: Time_CGE_RGE}
\end{wrapfigure}
We find that {\CGE} also has a \textit{computatoin efficiency merit} over {\RGE} when training CNNs in \textbf{Fig.\,\ref{fig: Acc_CGE_RGE}}.
\textbf{Fig.\,\ref{fig: Time_CGE_RGE}} presents the computation time of different training methods against the model size and highlights that {\CGE} is more computationally efficient than {\RGE}. 
To  examine the efficiency merit of {\CGE} in detail, we dissect the computation of a single ZO gradient estimate   into \textit{four stages}: \ding{172} generation of directional vectors, \textit{i.e.}, $\mathbf u_i$ or $\mathbf e_i$ in \eqref{eq: RGE_CGE}, \ding{173} perturbation of model weights, \textit{i.e.}, $\btheta + \mu \mathbf u_i$ or $\btheta + \mu \mathbf e_i$, 
\ding{174} model inference for calculating finite differences, and \ding{175} other arithmetic operations for gradient estimation. 
Our results show that {\RGE} takes much longer 
than {\CGE} in stages \ding{172} and \ding{173}. 
This is because, at each model query, {\RGE} needs to generate a directional perturbation vector with the same dimension as $\btheta$ 
and apply this perturbation to $\btheta$. 
By contrast, {\CGE} can generate the basis vector for `free' and perturb the model weights using a coordinate-wise indexing operation. 
\textbf{Tab.\,\ref{tab: rge_cge_time_compare}} 
shows a comparison of {\RGE} and {\CGE} in \ding{172}-\ding{175}.
It is also worth noting that a concurrent study \citep{malladi2023fine} argues that language model fine-tuning using RGE only necessitates the query budget $q=1$. However, it is important to highlight that their approach has certain restrictions, as it relies on having a high-quality pretraining model and a meticulously crafted prompt.

\begin{wraptable}{r}{65mm}
\vspace*{-4mm}
\centering
\caption{\footnotesize{
 Average per-iteration computation time comparison (seconds) between \txtr{RGE} and \txtg{CGE}. Here the gradient estimation process is dissected into 4 stages: \ding{172} generation of \underline{d}irectional \underline{v}ectors (DV), \ding{173}   model \underline{w}eights \underline{p}erturbation (WP), 
\ding{174} \underline{m}odel \underline{i}nference (MI) for calculating finite differences, and \ding{175} other \underline{a}rithmetic \underline{o}perations (AO) for gradient estimation.
}}
\label{tab: rge_cge_time_compare}
\vspace*{1mm}
\resizebox{65mm}{!}{%
\begin{tabular}{c|c|c|c|c}
\toprule[1pt]
\midrule
\multirow{2}{*}{Parameter \#} & \multicolumn{4}{c}{\txtr{RGE} ($q=d$) / \txtg{CGE} }  \\
&&&& \\
                       ($d$)  & \ding{172} DV               & \ding{173} WP                        & \ding{174} MI                        & \ding{175} AO \\ \midrule
730 & \txtr{0.173} / \txtg{0} & \txtr{0.152} / \txtg{0.026} & \txtr{0.202} / \txtg{0.21} & \txtr{0.149} / \txtg{0.0312}  \\
3700 & \txtr{1.43} / \txtg{0} & \txtr{1.2} / \txtg{0.0907} & \txtr{2.11} / \txtg{1.98} & \txtr{1.1} / \txtg{0.124} \\
6670 & \txtr{3.49}	/ \txtg{0} & \txtr{2.93} / \txtg{0.159}	& \txtr{4.84} / \txtg{4.53}  & \txtr{2.68} / \txtg{0.225}  \\
9640 &	\txtr{6.74} / \txtg{0}	& \txtr{5.47} / \txtg{0.246} &	\txtr{8.3} / \txtg{8.09}	& \txtr{4.97} / \txtg{0.336}	  \\
12610 &	\txtr{10.2} / \txtg{0} &	\txtr{8.86} / \txtg{0.457} &	\txtr{13.1} / \txtg{14.6} &	\txtr{8.05} / \txtg{0.566} \\ \midrule
\bottomrule[1pt]
\end{tabular}%
}
\end{wraptable}
\vspace*{-1mm}
\textbf{Tab.\,\ref{tab: rge_cge_time_compare}} provides a comparison of the average computation time per iteration for RGE and CGE-based ZO training across four stages: 
\ding{172} generation of directional vectors, \textit{i.e.}, $\mathbf u_i$ or $\mathbf e_i$ in \eqref{eq: RGE_CGE}, \ding{173} perturbation of model weights, \textit{i.e.}, $\btheta + \mu \mathbf u_i$ or $\btheta + \mu \mathbf e_i$, 
\ding{174} model inference for calculating finite differences, and \ding{175} other arithmetic operations for gradient estimation. 
 The comparison is conducted for training models of different numbers of parameters ($d$), ranging from $730$ to $12610$. 
In stage \ding{172},  RGE incurs computation time due to the generation of random Gaussian vectors, whereas CGE is free of this step since it estimates gradients in a coordinate-wise manner. In  stage \ding{173}, RGE requires more computation time compared to CGE across all parameter sizes. This is because RGE perturbs all model weights in a single query, while CGE only perturbs a single model weight coordinate per query.
In  stage \ding{174}, the computation times for both ZO algorithms increase with the number of parameters, with similar time consumption. 
In stage \ding{175},
RGE continues to require more computation time than CGE for all parameter sizes. This is because RGE needs to perform extra  arithmetic operations (such as the sum operation) to aggregate multiple queries, whereas CGE handles multiple queries coordinate-wise, resulting in a more efficient computation.

\section{Performance of Model Pruning via {\zograsp} }
\label{app: pruning}
We implement {\zograsp} using {\RGE} with the query number   $q=192$ (smaller than the model size $d$) and compare its performance with random pruning and FO-{\grasp}. 
All {\grasp} variants are performed over randomly initialized model parameters ($\btheta$).  
\textbf{Tab.\,\ref{tab: pruning_com_rn20}} and \textbf{Tab.\,\ref{tab: pruning_com_rn18}} present the pruning performance vs. model pruning ratios in the data-model setup of (CIFAR-10, ResNet-20) and (CIFAR-10, ResNet-18), respectively. 
Here the pruning performance is evaluated by measuring the testing accuracy of a sparse model on CIFAR-10. To assess the impact of model pruning, we utilize FO SGD as the optimizing method for sparse model training.
As we can see, our proposed ZO-GraSP method demonstrates comparable testing accuracy to FO-GraSP across various pruning ratios and significantly outperforms random pruning. 

\begin{table}[htb]
\centering
\caption{\footnotesize{Performance comparison of different model pruning methods on (CIFAR-10, ResNet-20). 
}}
\resizebox{\textwidth}{!}{%
\begin{tabular}{c|cccccccccc}
\toprule[1pt]
\midrule
 &  \multicolumn{10}{c}{(CIFAR-10, ResNet-20)} \\
 Pruning ratio  & $10\%$ & $20\%$ & $30\%$ & $40\%$ & $50\%$ & $60\%$ & $70\%$ & $80\%$ & $90\%$ & $95\%$ \\
\midrule
 \textbf{Random} & 92.71 $\pm$ 0.18 & 92.63 $\pm$ 0.04 & 92.40 $\pm$ 0.14 & 91.94 $\pm$ 0.03 & 91.77 $\pm$ 0.16 & 91.41 $\pm$ 0.20 & 90.57 $\pm$ 0.05 & 89.35 $\pm$ 0.19 & 86.55 $\pm$ 0.15 & 82.51 $\pm$ 0.16 \\
\midrule
 \textbf{FO-GraSP} & 92.64 $\pm$ 0.10 & 92.44 $\pm$ 0.06 & 92.34 $\pm$ 0.11 & 92.35 $\pm$ 0.14 & 92.07 $\pm$ 0.11 & 92.00 $\pm$ 0.03 & 91.63 $\pm$ 0.16 & 90.75 $\pm$ 0.07 & 88.55 $\pm$ 0.09 & 85.48 $\pm$ 0.26 \\
\midrule
\textbf{{\zograsp}} & 92.74 $\pm$ 0.07 & 92.56 $\pm$ 0.10 & 92.58 $\pm$ 0.20 & 92.46 $\pm$ 0.05 & 92.09 $\pm$ 0.10 & 91.61 $\pm$ 0.13 & 91.48 $\pm$ 0.23 & 90.21 $\pm$ 0.16 & 88.08 $\pm$ 0.09 & 84.80 $\pm$ 0.10 \\
\midrule
\bottomrule[1pt]
\end{tabular}%
}
\label{tab: pruning_com_rn20}
\vspace*{-4mm}
\end{table}

\begin{table}[ht]
\centering
\caption{\footnotesize{Performance comparison of different model pruning methods on (CIFAR-10, ResNet-18).}}
\resizebox{\textwidth}{!}{%
\begin{tabular}{c|cccccccccc}
\toprule[1pt]
\midrule
   & \multicolumn{10}{c}{(CIFAR-10, ResNet-18)} \\
Pruning ratio   & $10\%$ & $20\%$ & $30\%$ & $40\%$ & $50\%$ & $60\%$ & $70\%$ & $80\%$ & $90\%$ & $95\%$ \\
\midrule
 \textbf{Random}& 95.45$\pm$0.06 & 95.51$\pm$0.11 & 95.32$\pm$0.15 & 95.34$\pm$0.10 & 95.40$\pm$0.20 & 95.09$\pm$0.13 & 94.92$\pm$0.14 & 94.58$\pm$0.06 & 93.53$\pm$0.10 & 92.13$\pm$0.26 \\
\midrule
 \textbf{{FO-GraSP}} & 95.44$\pm$0.07 & 95.46$\pm$0.30 & 95.53$\pm$0.06 & 95.49$\pm$0.05 & 95.44$\pm$0.08 & 95.86$\pm$0.09 & 95.99$\pm$0.17 & 95.97$\pm$0.05 & 95.91$\pm$0.03 & 96.19$\pm$0.10 \\
\midrule
\textbf{{\zograsp}} & 95.43$\pm$0.05 & 95.53$\pm$0.03 & 95.53$\pm$0.14 & 95.56$\pm$0.18 & 95.50$\pm$0.06 & 95.24$\pm$0.09 & 95.36$\pm$0.05 & 95.28$\pm$0.05 & 94.94$\pm$0.08 & 94.39$\pm$0.13 \\
\midrule
\bottomrule[1pt]
\end{tabular}%
}
\label{tab: pruning_com_rn18}
\vspace*{-4mm}
\end{table}

\section{Algorithm Details}
\label{app: alg_details}

\begin{algorithm}[H]
  \caption{{\zograsp}-oriented-LPR-guided ZO training}
  \label{alg: lpr_zo}
  \begin{algorithmic}[1]
  \State Get $\mathcal S_{\zograsp}$ through {\zograsp}  
  \eqref{eq: grad_vec_product_approx1}
  \State Obtain layer-wise pruning ratio $\mathcal S_{\text{layer}}$ based on $\mathcal S_{\zograsp}$
    \For{Epoch $t = 0,1,2, \ldots, T-1$} 
      \State Randomly generate a sparse coordinate set $\mathcal{S}_t$ according to $\mathcal S_{\text{layer}}$
      \For{Iterations per epoch} 
        \State Obtain \eqref{eq: CGE_sparsity} $  \hat \nabla_{\boldsymbol \theta}   \ell ({\boldsymbol \theta} ) $ based on $\mathcal{S}_t$ 
        \State Update model weights: $\boldsymbol{\theta} \leftarrow \btheta -    \alpha   \hat \nabla_{\boldsymbol \theta}   \ell ({\boldsymbol \theta} )  $
        \EndFor
    \EndFor
  \end{algorithmic}
\end{algorithm}

\paragraph{{\zograsp}-oriented-LPR-guided ZO training}   \textbf{Algorithm\,\ref{alg: lpr_zo}} shows the algorithmic steps to fulfill the {\zograsp}-oriented-LPR-guided ZO training. At initialization, the algorithm acquires the coordinate set of unpruned model weights
$\mathcal S_{\zograsp}$ by applying our proposed ZO-GraSP as shown in \eqref{eq: grad_vec_product_approx1}. Subsequently, it computes the layer-wise pruning ratios, denoted as $\mathcal S_{\text{layer}}$, derived from $\mathcal S_{\zograsp}$. 
The \eqref{eq: CGE_sparsity}  $  \hat \nabla_{\boldsymbol \theta}   \ell ({\boldsymbol \theta} ) $ is then estimated using
 $\mathcal S_{\text{layer}}$.
The model weights are then updated by subtracting the estimated sparse gradient to the current weight $\boldsymbol{\theta}$. This process iteratively refines the weights for optimizing the model's performance.

\begin{wrapfigure}{r}{49mm}
\vspace*{-5mm}
\centerline{
\includegraphics[width=49mm,height=!]{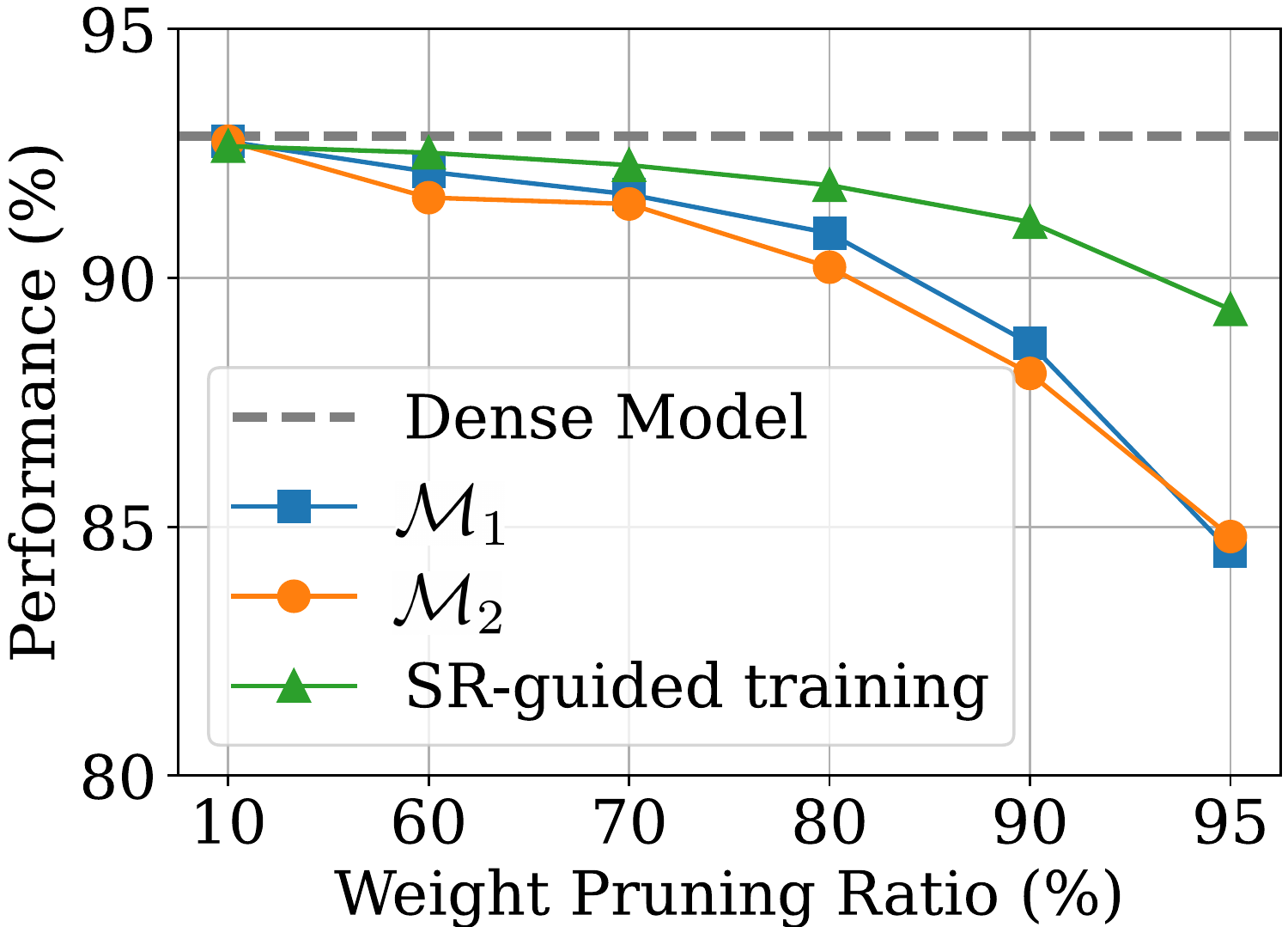}
}
\caption{\footnotesize{
Performance (testing accuracy)   of SR-guided training vs. pruning ratios on (CIFAR-10, ResNet20).
}}
  \label{fig: sparse_training_SR}
\end{wrapfigure}
\textbf{Fig.\,\ref{fig: sparse_training_SR}} demonstrates  the superiority of SR-guided training to another two alternative methods, {$\mathcal M_1$} built on  alternative optimization between {\zograsp} and sparse training, and 
$\mathcal M_2$ based on {pruning before model training}. 
To examine the effectiveness of  SR in integrating  {\zograsp} with model training, we compare the pruning performance in the FO training regime (\textit{i.e.}, the resulting sparse model is trained using FO methods).  As we can see, SR-guided training consistently outperforms methods {$\mathcal M_1$} and {$\mathcal M_2$}, with the accuracy improvement becoming more pronounced as the pruning ratio increases. As the pruning ratio increases, the performance gap with the original dense model also widens. However, our method achieves the smallest performance gap compared to the FO baseline.

\paragraph{DeepZero framework.}
We introduce the DeepZero framework, a novel approach designed to train models using ZO optimization. This framework assimilates three key features: the implementation of {\zograsp}-oriented-LPR-guided ZO training, the reuse of features, and the adoption of parallel computation techniques. These components of the DeepZero framework are expounded in \textbf{Algorithm\,\ref{alg: deepzo}}.

\begin{algorithm}[H]
  \caption{DeepZero Framework 
  }
  \label{alg: deepzo}
  \begin{algorithmic}[1]
  \State \textbf{Initialize:} Total epochs $T$, sparsity update interval $K_\mathrm{sparse}$, LPRs $\mathcal{R}$ via ZO-GraSP \eqref{eq: grad_vec_product_approx1}
  \For {Epoch $t = 0, 1, 2, \ldots, T-1$ 
  }
  
  \If {$t$ \texttt{mod} $K_\mathrm{sparse}==0$} \hfill \Comment{\textcolor{blue}{Sparsity inducing}}
  \State Update sparse coordinate index set  $\mathcal S_{t}$ according to $\mathcal{R}$
  \EndIf
    \For{Process $i = 1,2, \ldots, M$} \hfill \Comment{\textcolor{blue}{$M$ processes}}
    \State \parbox[t]{\dimexpr\textwidth-\leftmargin-\labelsep*2-\labelwidth*2}{Based on $\mathcal S_{t}$, parallelized forward  evaluations with feature reuse \eqref{eq: f_network_layers}-\eqref{eq: Parallel_CGE}}
    \EndFor
    \State \eqref{eq: CGE_sparsity} estimation using above model evaluations
    \State Model weights updating and synchronization
  \EndFor
  \end{algorithmic}
\end{algorithm}

\section{Additional Experiments of Image Classification}
\label{app: res20_exp}

\begin{figure}[htb]
    \begin{minipage}[t]{0.308\textwidth}
    \centering
    \includegraphics[width=\textwidth]{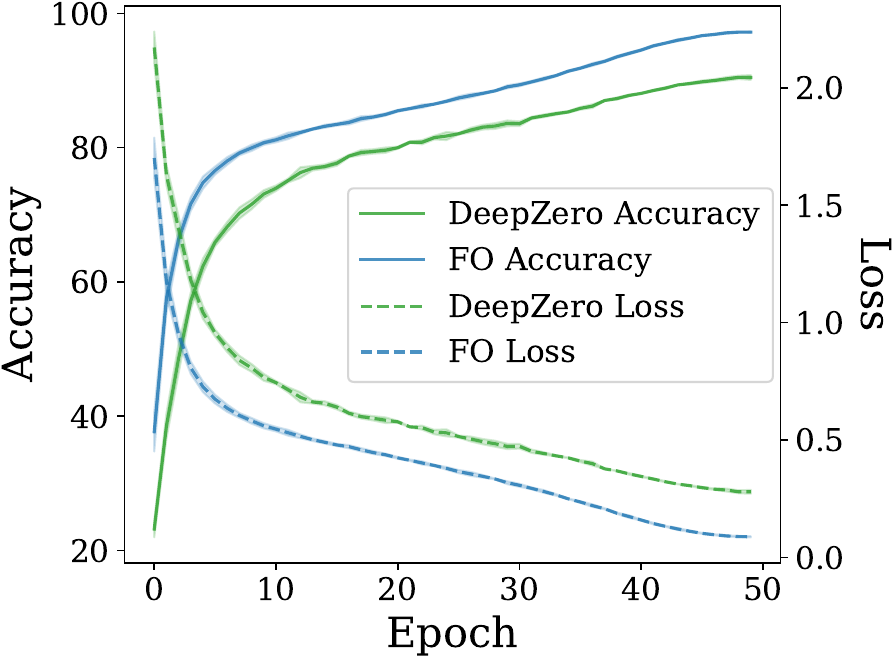}
    \caption{\footnotesize{Training trajectory of {\DeepZero} and FO training on (CIFAR-10, ResNet-20).   {\DeepZero} adopts \ref{eq: CGE_sparsity} of   $90\%$ sparsity. The mean and standard deviation of 3 independent runs are reported.}}
    \label{fig: training_traj}
    \end{minipage}\hfill
    \begin{minipage}[t]{0.344\textwidth}
    \centering
    \includegraphics[width=\textwidth]{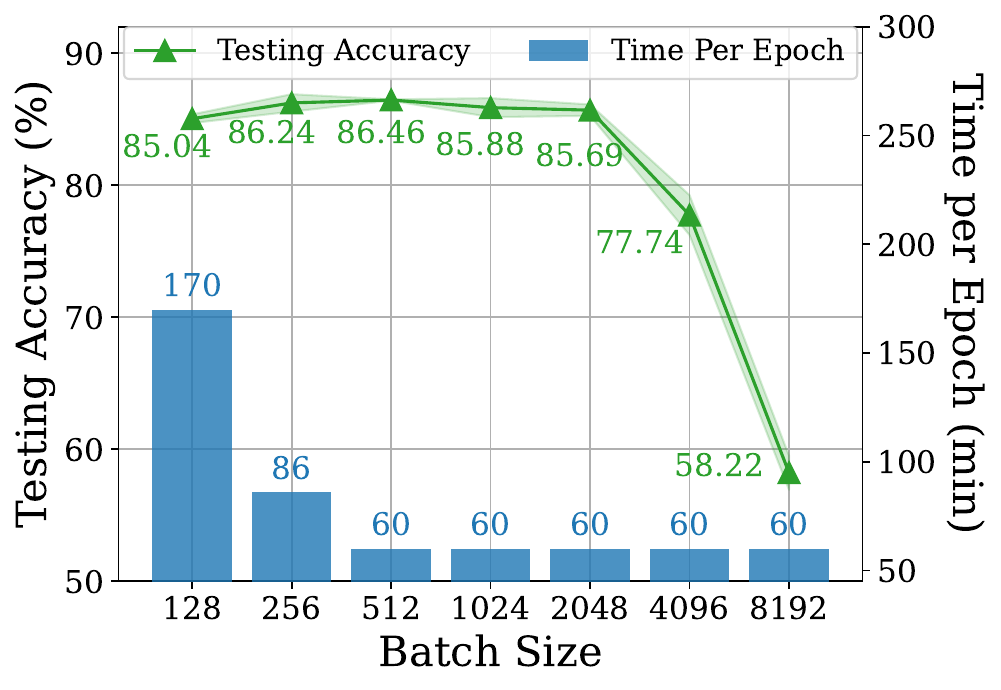}
    \caption{\footnotesize{Accuracy performance and computation time of 
    {\DeepZero}  on (CIFAR-10, ResNet-20) against different batch sizes. Other settings are consistent with Fig.\,\ref{fig: training_traj}.
    }}
    \label{fig: batch_size}
    \end{minipage}\hfill
    \begin{minipage}[t]{0.308\textwidth}
    \centering
    \includegraphics[width=\textwidth]{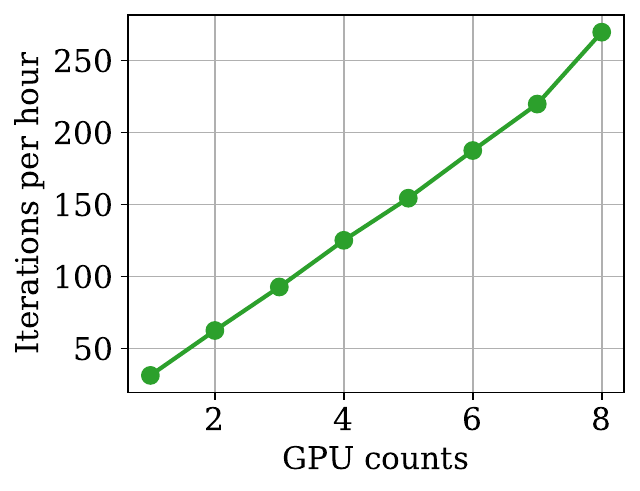}
    \caption{\footnotesize{The training speed of
    {\DeepZero}  on (CIFAR-10, ResNet-20) against GPU counts. Experiments are conducted under    NVIDIA TESLA V100 16G GPU resources. }}
    \label{fig: gpu_number}
    \end{minipage}
\end{figure}

\textbf{Fig.\,\ref{fig: training_traj}} presents the training trajectory comparison between {\DeepZero} and the dense FO training baseline. We observe that despite the marginal performance drop, {\DeepZero} achieves a competitive convergence rate as the FO counterpart. Besides, it is important to highlight that due to the small variance of CGE on a reduced dimension (parameters after pruning), {\DeepZero} achieves competitively stable training as the FO baseline, as indicated by similar standard deviations.

\textbf{Fig.\,\ref{fig: batch_size}} presents the performance and the time consumption of {\DeepZero} in various batch size settings. The impact of batch size on model performance has been well-documented in FO training, with larger batch sizes often leading to decreased performance.
We sought to examine the effect of batch size selection on ZO training and determine whether increasing the size could reduce training time.
As depicted in \textbf{Fig.\,\ref{fig: batch_size}}, we observed a marked decline in performance when the batch size exceeded 2048. Additionally, when the batch size was set to 512, it reached a constant time consumption of 60 minutes per epoch, rendering any further reduction in training time infeasible due to the limitations of GPU computational capacity.
We stress that reaching this constant time consumption signs the full utilization of each GPU, which is easily achieved by forward parallelization.

\textbf{Fig.\,\ref{fig: gpu_number}} presents {\DeepZero}'s training speed (iterations per hour) on CIFAR-10 and ResNet-20 when using different counts of GPUs. We observe training speed scaling linearly with regard to the GPU counts, which justifies the efficiency of acceleration by forward pass and feature reuse. 

\vspace{-5mm}
\section{Additional Illustration of Black-box Defense against Adversarial Attacks}
\label{app: black_box_defense}

\begin{wrapfigure}{l}{53mm}
 \vspace*{-4.8mm}
\centerline{
\includegraphics[width=53mm,height=!]{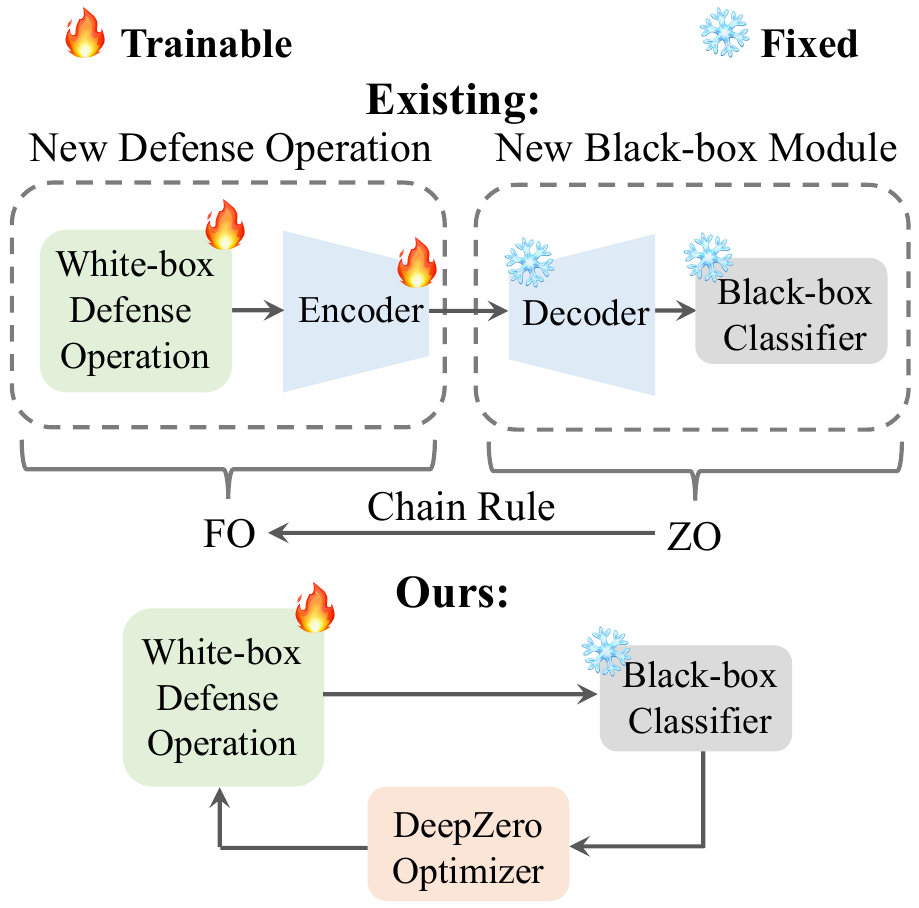}
}
\caption{\footnotesize{Schematic overview of baseline ZO-AE-DS \citep{zhang2022robustify} and DeepZero-enabled black-box defense. 
}}
  \label{fig: ae_ds}
 \vspace*{-4.8mm}
\end{wrapfigure}
In \textbf{Fig.\,\ref{fig: ae_ds}}, we compare our method, DeepZero, with the baseline method ZO-AE-DS \citep{zhang2022robustify}. ZO-AE-DS utilizes an autoencoder (AE) to reduce the dimensionality of the ZO gradient estimation by merging the AE's decoder with the black-box classifier and integrating the AE's encoder with the white-box defense operation. However, ZO-AE-DS suffers from poor scalability to high-resolution datasets like ImageNet due to the use of AE, which compromises the fidelity of the image input to the black-box classifier. In contrast, DeepZero directly optimizes the white-box defense operation without the need for an AE, as demonstrated in {Fig.\,\ref{fig: ae_ds}}.

Following \citep{salman2020denoised, zhang2022robustify}, we model the defense operation as a denoiser given by  DnCNN \citep{zhang2017beyond}. To optimize it for black-box defense, we adopt the `pretraining + fine-tuning' approach, as detailed in \citep{zhang2022robustify}. We  first employ the Adam optimizer \citep{kingma2014adam} to pretrain the denoiser over 90 epochs, based on the mean squared error (MSE) loss function. The pre-training stage excludes the optimization with the  black-box image classifier. 
After the pretraining phase, we apply ZO-GraSP to the pretrained denoiser, with  the pruning ratio of $95\%$. Notably, the percentage of unpruned parameters    of the denoiser tends to be almost zero for all layers, except the first and the last layers. 
With this insight in mind, we specify the finetuning strategy as partial finetuning, which  targets only the last layer of the denoiser. However, we still need to address the black-box optimization challenge, provided by the black-box classifier $f(\mathbf x)$. Let $\btheta$ represent the denoiser to be finetuned, DeepZero then solves the following black-box optimization problem:
\begin{align}
  \hspace*{-2.5mm}  \begin{array}{ll}
    \displaystyle \minimize_{\boldsymbol \theta }     &   \hspace*{-1.5mm} 
  \mathbb E_{\bdelta, \mathbf x}  [  \ell_{\mathrm{CE}} (f ( D_{\boldsymbol \theta}(\mathbf x+\bdelta) ), f    (\mathbf x)   ) ],
    \end{array}
    \label{eq: DS}
\end{align}
where $\mathbf x$ signifies the input, $\ell_{\mathrm{CE}}$ refers to the Cross-Entropy (CE) loss, and $\boldsymbol \delta \in \mathcal N(\mathbf 0, \sigma^2 \mathbf I)$ represents the standard Gaussian noise with variance $\sigma^2$.
We apply DeepZero to solve problem \eqref{eq: DS} under 20 training epochs, and use a learning rate of $10^{-5}$, which is reduced by a factor of $10$ every four epochs.

\section{Simulation-coupled DL for Discretized PDE Error Correction}
\label{app: sitl}

\begin{wrapfigure}{r}{63mm}
\vspace*{-3.8mm}
\centerline{
\includegraphics[width=63mm,height=!]{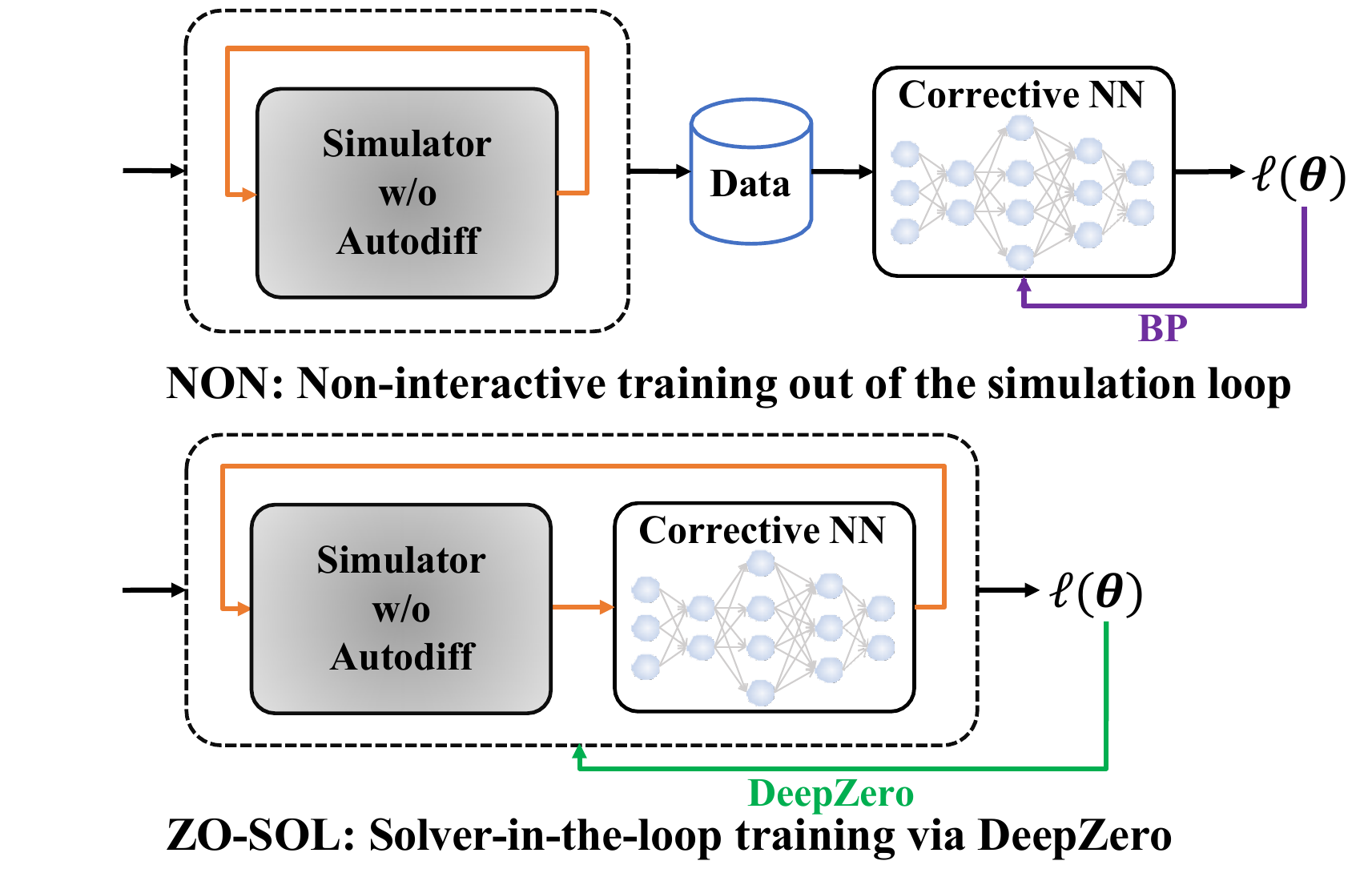}
}
\caption{\footnotesize{
Schematic overview of the baseline approach NON given by training over pre-generated simulation data and ZO-SOL by leveraging DeepZero to address the solver-in-the-loop training challenge.
}}
  \label{fig: sitl_methods_illustration}
 \vspace*{-3.8mm}
\end{wrapfigure}
The feasibility of training a corrective NN through looping interactions with the iterative partial differential equation (PDE) solver, coined `solver-in-the-loop' ({SOL}), has been demonstrated in \citep{um2020solver}. 
By leveraging {\DeepZero}, we can enable the use of SOL with non-differentiable, or black-box, simulators \cite{hu2019chainqueen, fang2022complex}. 
We name our method \textbf{ZO-SOL}. \textbf{Fig.\,\ref{fig: sitl_methods_illustration}} presents a comparison with the baseline method NON, given by non-interactive training out of the simulation loop using pre-generated low and high-fidelity simulation data.

In our experiments, we consider an unsteady wake flow simulation—a standard benchmark case for fluid dynamics \citep{um2020solver}—to assess the efficacy of our method. The corrective neural network, adopted from \citep{um2020solver}, is a sequential convolutional network consisting of two convolutional layers with $5 \times 5$ kernels, totaling 56,898 parameters. The SOL problem can then be formulated  as the optimization problem
\begin{align}
  \hspace*{-2.5mm}  \begin{array}{ll}
    \displaystyle \argmin_{\boldsymbol \theta }   
    & \sum_{t=0}^{T-n} \sum_{i=0}^{n-1}\|\mathcal{P}_{{s}}(\Tilde{\bm s}_{t+i})+\mathcal{C}_{\btheta}(\mathcal{P}_{{s}}(\Tilde{\bm s}_{t+i}))- \mathbf y_{t+i+1}\|^2 
    \end{array},
    \label{eq: sol}
\end{align}
where $\mathcal{P}_{{s}}$ denotes the 
PDE solver used for the low-fidelity simulation, $\mathcal{C}_{\btheta}$ signifies the corrective neural network parameterized by $\btheta$,   $\mathbf y_{t+i}$ stands for the high-precision simulation state (regarded as the ground truth) at time $t+i$, $T$ is the number of timesteps for the ground truth simulation, $n$ is the number of unrolling steps used in SOL, 
and $\Tilde{\bm s}_{t+i} $ signifies the output from  SOL at the $i$th step, namely, 
\begin{align*}
\Tilde{\bm s}_{t+i} = 
\underbrace{
[(\mathds{1}+\mathcal{C}_{\btheta})\circ \mathcal{P}{{s}}] \circ [(\mathds{1}+\mathcal{C}_{\btheta})\circ \mathcal{P}{{s}}] \circ 
\cdots \circ
[(\mathds{1}+\mathcal{C}_{\btheta})\circ \mathcal{P}{{s}}]
}_{i~\mathrm{times}}
(\mathbf y_t),
\end{align*}
where $\circ$ is the function composition operation, $\mathds{1}$ is an identity function and $(\mathds{1}+\mathcal{C}_{\btheta})$ stands for a function expressed as $(\mathds{1}(\cdot)+\mathcal{C}_{\btheta}(\cdot))$.
Note that in our experiments, we take $n=16$ during training, that is we have 16 unrolling steps. 

For the unsteady wake flow simulation, we follow the details outlined in \citep{um2020solver} which we include here for completeness. 
Namely, the unsteady wake flow simulation solves a discretized version the the incompressible Navier-Stokes equations given by:
\begin{align}
    \frac{\partial u_x}{\partial t} + \bm{u} \nabla u_x &= - \frac{1}{\rho} \nabla p  + \nu \nabla \cdot \nabla u_x \label{eq:navier-1} \\
    \frac{\partial u_y}{\partial t} + \bm{u} \nabla u_y &= - \frac{1}{\rho} \nabla p  + \nu \nabla \cdot \nabla u_y \label{eq:navier-2} \\
    \nabla \cdot \bm{u} &= 0, \label{eq:navier-3}
\end{align}
where $t$ is time, $\rho$ is density, $\nu$ is viscosity, $p$ is pressure, $\bm{u} = (u_x, u_y)^T$ is velocity ($x$ and $y$ directions), and $\nabla \cdot$ denotes the divergence operator.
The domain is $\Omega = [0,1] \times [0,2]$ with open boundary conditions and an initial state of $\bm{u} = (0,1)^T$ along $x \in [0,1]$ and $y = 0$.
Additionally, there is a circular rod of diameter $0.1$ at position $(0.5, 0.5)^T \in \Omega$.
The domain $\Omega$ for the low and high fidelity simulations is discretized using a staggered grid of dimensions $[32, 64]$ and $[64,128]$, respectively. 
Additionally, we take $T=500$ timesteps. 
We utilized the PhiFlow \citep{holl2020learning} simulation code as the solver for our experiments and we based our implementation off of the solver in the loop training process 
from the code \url{https://github.com/tum-pbs/Solver-in-the-Loop}.

For simplicity, the loss in (\ref{eq: sol}) is only expressed for a single simulation with $T$ timesteps. 
In our experiments, we utilize 6 different simulations in the training dataset each of which have different Reynolds numbers (which affects fluid flow turbulence and can be altered by varying $\nu$ in (\ref{eq:navier-1}) and (\ref{eq:navier-2})) in the set $\{ 97.7, 195.3, 390.6, 781.3, 1562.5, 3125.0 \}$, consistent with \citep{um2020solver}.
Hence, during training the loss function has an additional outer summation to account for the 6 different training simulations.
The test set consists of five different simulations computed using Reynolds numbers in the set $\{ 146.5, 293.0, 585.9, 1171.9, 2343.8 \}$, also consistent with \citep{um2020solver}.

To solve problem \eqref{eq: sol}, DeepZero under the Adam optimization framework is used with a learning rate of $10^{-4}$ over ten training epochs.

\section{Broader Impacts and Limitations}
\label{app: broader_impact}

\paragraph{Broader Impacts.} Our proposed ZO learning for deep model training offers a valuable solution for various ML problems that involve interactions with black-box APIs, such as language models as a service, and on-chip learning problems where gradient calculation is infeasible on resource-limited hardware. The applications of DeepZero and its black-box approaches explored in this work can also contribute to advancements in optimization theory and model pruning in other fields. The insights gained from studying DeepZero's scalability, efficiency, and effectiveness can have far-reaching implications beyond the realm of deep learning.

\paragraph{Limitations.} One limitation of our approaches is the high number of model queries required, which is inherent to ZO optimization in general. Improving query efficiency is an important area for future research and compute-efficient techniques from \citep{bartoldson2023compute} likely will be helpful. Additionally, the infusion of sparse deep priors in ZO training may not be suitable for non-deep learning models. Therefore, it remains crucial to develop more generic and model-agnostic ZO training methods that can handle large-scale and extreme-scale problems effectively.

\end{document}